\documentclass[10pt]{article}

\makeatletter
\newcommand*{\addFileDependency}[1]{
  \typeout{(#1)}
  \@addtofilelist{#1}
  \IfFileExists{#1}{}{\typeout{No file #1.}}
}
\makeatother

\usepackage{float}
\usepackage[margin=1in]{geometry}
\usepackage{amsmath,amsthm,amssymb}
\usepackage{graphicx}
\usepackage{titlesec}
\usepackage{natbib}
\usepackage{authblk}
\usepackage{caption}
\usepackage{subfig}
\usepackage{textcomp}
\usepackage{enumerate}
\usepackage{tabularx,ragged2e}
\usepackage{multirow}
\usepackage{color}
\usepackage{diagbox}
\usepackage[utf8]{inputenc}
\usepackage{threeparttable}
\usepackage{tabularx,ragged2e}
\usepackage{txfonts}
\usepackage{caption}
\usepackage{booktabs}
\usepackage{stfloats}
\usepackage{makecell}
\usepackage{multicol}
\definecolor{red}{rgb}{1,0,0}
\definecolor{blue}{rgb}{0,0,1}
\usepackage{algorithm}
\usepackage{algorithmicx}
\usepackage{algpseudocode}  
\usepackage{amsfonts}
\usepackage{flushend}
\usepackage{balance}
\usepackage{longtable}   
\usepackage{bm}
\usepackage{url}

\usepackage{hyperref}
\usepackage{xr-hyper}
\usepackage[table]{xcolor}

\definecolor{group1}{RGB}{228, 236, 255}    
\definecolor{group2}{RGB}{255, 242, 230}    
\definecolor{group3}{RGB}{240, 248, 245}    

\definecolor{group1dark}{RGB}{200, 220, 255} 
\definecolor{group2dark}{RGB}{255, 230, 200} 
\definecolor{group3dark}{RGB}{222, 238, 212} 

\newcommand{\del}[1]{}

\renewcommand\baselinestretch{2}

\titleformat*{\section}{\large\bfseries}
\titleformat*{\subsection}{\normalsize\bfseries}
\titleformat*{\subsubsection}{\small\bfseries}

\begin{document}

\title{Resolving compositional and conformational heterogeneity in cryo-EM with deformable 3D Gaussian representations}
\renewcommand\footnotemark{}
\author{Bintao He$^{1,\dagger}$, Yiran Cheng$^{1,2,\dagger}$, Hongjia Li$^{3}$, Xiang Gao$^{4}$, Xin Gao$^{5,*}$, Fa Zhang$^{3,*}$,  Renmin Han$^{1,2,*}$  \thanks{$^{\dagger}$These authors should be regarded as Joint First Authors; $^*$All correspondence should be addressed to Xin Gao (xin.gao@kaust.edu.sa), Fa Zhang (zhangfa@bit.edu.cn) and Renmin Han (hanrenmin@sdu.edu.cn).}}
\date{\small $^1$Research Center for Mathematics and Interdisciplinary Sciences (Ministry of Education Frontiers Science Center for Nonlinear Expectations), Shandong University, Qingdao 266237, China; $^2$College of Medical Information and Engineering, Ningxia Medical University, Yinchuan 750004, China; $^3$School of Medical Technology, Beijing Institute of Technology, Beijing 100081, China; $^4$State Key Laboratory of Microbial Technology, Shandong University, Qingdao 266237, China; $^5$King Abdullah University of Science and Technology (KAUST), Computational Bioscience Research Center (CBRC), Computer, Electrical and Mathematical Sciences and Engineering (CEMSE) Division, Thuwal, 23955, Saudi Arabia.}

\maketitle

\renewcommand\baselinestretch{2}

\vspace{-2em}
\begin{abstract}
Understanding protein flexibility and its dynamic interactions with other molecules is essential for studying protein function. Although cryogenic electron microscopy (cryo-EM) provides an opportunity to observe macromolecular dynamics directly, computational analysis of datasets mixing continuous and discrete structural states remains a formidable challenge. Here we introduce GaussianEM, a Gaussian-based pseudo‑atomic framework that simultaneously resolves compositional and conformational heterogeneity from cryo‑EM images. GaussianEM employs a dual-encoder–single-decoder architecture to decompose images into learnable Gaussian components, with variability encoded through modulated parameters. This explicit parameterization yields a continuous, intuitive representation of conformational dynamics that inherently preserves local structural integrity. By modeling displacements in Gaussian space, we capture atomic‑scale conformational landscapes, bridging density maps and all‑atom models. In comprehensive experiments, GaussianEM successfully reconstructs complex compositional and conformational variability, and resolves previously unobserved details in public datasets. Quantitative evaluations further confirm its ability to capture broader conformational diversity without sacrificing structural fidelity. 
\end{abstract}

\section{Introduction}
Cryogenic electron microscopy (cryo-EM) is a leading technique for imaging biological macromolecules at near-atomic resolution. The structural variability of these molecules is fundamental to their biological functions. In single particle analysis, plunge-frozen images of numerous individual complexes contain rich molecular information that reflects their compositional and/or conformational heterogeneity \cite[]{lyumkis2019challenges, glaeser2016good, tang2023conformational}. However, the high level of experimental noise makes it difficult to extract meaningful structural information from a single snapshot. To address this, averaging is typically combined with classification to divide the cryo-EM dataset into multiple structurally similar subsets, enabling the reconstruction of stable and reliable three-dimensional structures \cite[]{scheres2016processing, scheres2012relion, punjani2017cryosparc, grant2018cis}. These approaches have proven effective for datasets exhibiting a discrete number of compositional variations, where each class contains a sufficient number of particles for reconstruction. Multi-body refinement \cite[]{nakane2018characterisation} further improves local variation estimation by dividing complexes into independently rigid-moving components \cite[]{ilca2015localized, xu2021conformational, holvec2024structure}. However, the limitation on the size of the independent parts restricts its application to highly flexible datasets or motions of small subunits.

Recently, advanced neural networks have been introduced to uncover detailed and complex structural variations in cryo-EM data. CryoDRGN \cite[]{zhong2021cryodrgn} was the first to map projection images to a continuous latent space using a neural radiance field (NeRF) framework. It thereby eliminates the need for user-defined discrete structural classes. Given particle poses obtained from homogeneous reconstruction, CryoDRGN bridges experimental images and reconstructed volumes in the Fourier domain according to the Fourier-slice theorem. By traversing the continuous latent space, macromolecular movies can be generated to visualize protein motions and capture both compositional and conformational heterogeneity. Its results demonstrate expressive potential for estimating structural variation from a single cryo-EM dataset. Building upon the NeRF architecture, several subsequent approaches \cite[]{luo2023opus, herreros2025real} represent 3D volumes directly in real space to achieve higher-resolution reconstructions. These methods also design different disentanglement strategies for properties such as particle poses and contrast transfer function (CTF) parameters, aiming to obtain a smoother and more physically meaningful latent space. CryoSTAR \cite[]{li2024cryostar} further incorporates structural priors and constraints from protein atomic models to enhance reconstruction quality and regulate the variation scale within the estimated conformational landscape. However, NeRF-based methods inherently require explicit 3D volume reconstruction for map comparison and are unable to directly visualize continuous structural transformations. Therefore, interpreting multiple points across the estimated conformation landscape is often time-consuming and relies heavily on manual inspection. In particular, local structures often lack consistency within the reconstructed serial volumes.

e2gmm \cite[]{chen2021deep} in EMAN2, represents the protein structure using a Gaussian mixture model (GMM) and intuitively describes structural heterogeneity through variations in Gaussian parameters. It measures the difference between experimental images and the GMM projection in terms of Fourier Ring Correlation (FRC) values, and constructs network inputs by stacking the FRC gradients with respect to the GMM. The explicit modeling of the existence and displacement of Gaussian spheres facilitates the localization of structural changes. However, the use of stacked gradient inputs constrains the number of Gaussians that can be handled for high-resolution reconstruction and reduces the model’s sensitivity to complex motions. DynaMight \cite[]{schwab2023dynamight} focuses on continuous motion estimation by optimizing Gaussian displacements in the Fourier domain. It additionally estimates inverse deformations to improve 3D consensus reconstruction and supports a larger number of Gaussian spheres derived from atomic models. 3DFlex \cite[]{punjani20233dflex} directly estimates a convection field to deform the 3D consensus structure, enabling the simultaneous generation of high-resolution 3D volumes and deformation fields corresponding to experimental images. However, both methods are limited in their ability to characterize compositional heterogeneity. Consequently, intuitive interpretation of both discrete and continuous conformational changes remains a challenge.

Here, we propose a cryo-EM heterogeneous reconstruction method, GaussianEM, to handle both compositional and conformational structural variations using Gaussian pseudo-atomic models. In GaussianEM, the macromolecular structure is represented as a set of 3D Gaussian functions (3D Gaussians), whose changes in presence and coordinates intuitively reflect local structural missing and motion, respectively. Unlike previous Gaussian-based approaches, GaussianEM can model tens of thousands of Gaussians at the pseudo-atomic level and jointly estimate their parameter variations. The framework employs a special two-encoder-one-decoder architecture: an image encoder maps individual experimental images into a continuous latent space, a Gaussian encoder assigns a unique identifier to each Gaussian, and a volume decoder predicts parameter variations to infer different conformations. The latent variable from the image encoder and the Gaussian embedding from the Gaussian encoder are combined to estimate the change associated with each Gaussian. In contrast to e2gmm, which reconstructs from projections, GaussianEM provides a direct volume reconstruction from 3D Gaussians. 
By applying deformations to a consensus model, GaussianEM ensures consistent structural changes in the local regions, an issue that may be observed in voxel-based methods. When an atomic model is available, GaussianEM naturally supports the translation of Gaussian variations into atomic conformational landscapes, thereby providing a direct bridge between density-based analysis and pseudo-atomic level interpretation. 


To evaluate the performance of GaussianEM, we conducted comprehensive experiments on multiple simulated benchmarks and EMPIAR datasets. The results demonstrate that GaussianEM effectively and intuitively captures both discrete and continuous conformational changes in the ribosome assembly, the pre-catalytic spliceosome,  the $\alpha V\beta8$ integrin, and the $\alpha V\beta8$ integrin bound to latent TGF-$\beta$.
Notably, GaussianEM identifies novel conformational states of T6SS effectors that were not resolved in previously published analyses.
Additionally, GaussianEM preserves the integrity and consistency of local structures during continuous transitions along the conformational trajectory, as it explicitly models deformations applied to the consensus Gaussian model. Quantitative evaluations focusing on conformational diversity demonstrate GaussianEM's distinct advantage over CryoDRGN, DynaMight, and 3DFlex in capturing a wider range of conformational states. Furthermore, FSC-based comparisons on benchmark datasets confirm the high fidelity and reliability of GaussianEM's reconstruction results.


\section{Results}

\subsection{The GaussianEM method}

\begin{figure}[H]
	\centering
	\includegraphics[width=0.9 \textwidth]{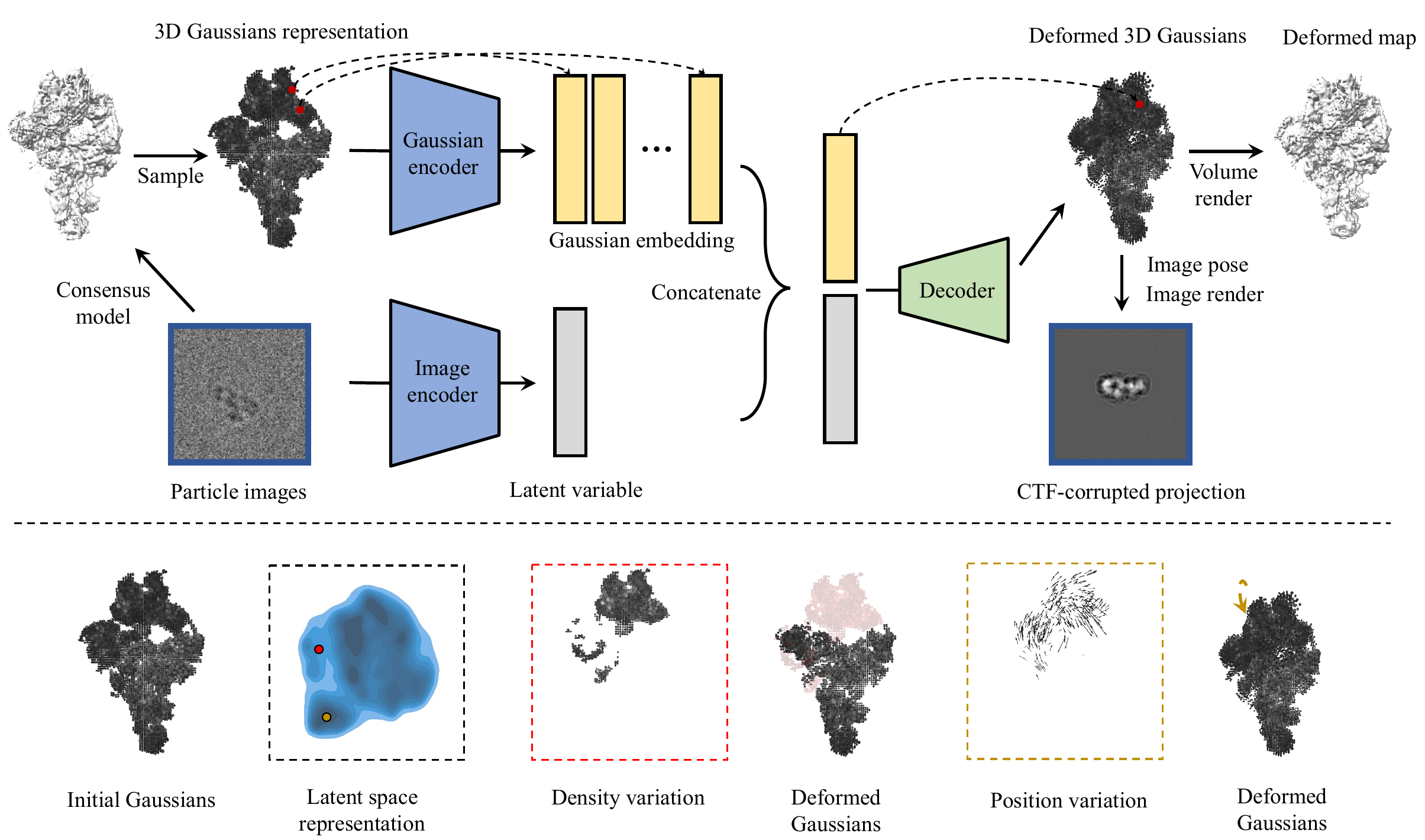}
	\caption{\textbf{Architecture of GaussianEM.} The GaussianEM method represents macromolecular structures using 3D Gaussians, and its heterogeneity analysis network consists of two encoders and one decoder. Each particle image is encoded into a continuous latent space, while each Gaussian is independently encoded to obtain a unique identifier (embedding). Then, the decoder combines the latent variable and each Gaussian embedding to estimate the parameter variations for Gaussians. The resulting deformed 3D Gaussians directly illustrate the structural changes of the protein, capturing both the compositional and conformational heterogeneity.}
	\label{fig:architecture}
\end{figure}

In GaussianEM, we assume that the heterogeneity present in experimental particle images can be represented by a single consensus structure along with its corresponding structural variations. Compositional heterogeneity is manifested as partial structural missing, whereas conformational heterogeneity appears as structural displacement. Figure \ref{fig:architecture}a illustrates the architecture of GaussianEM. The method uses a set of 3D Gaussians to fit the consensus model and identifies parameter variations of Gaussians by learning an adapted variation autoencoder (VAE). Each Gaussian is defined by seven parameters: a density value, a 3D scale vector used to compute its covariance, and 3D spatial coordinates. Local structural motions are described by changes in the 3D positions of Gaussians that make up the model, and the presence or absence of components is reflected by variations in their density values.
The initial Gaussians are uniformly sampled from the homogeneous reconstruction, filtered by the provided contour level. The number of Gaussians determines the achievable resolution of reconstructed structures, but also has a substantial impact on computational cost. In practice, GaussianEM adopts a sampling interval of two voxels, which works well on a wide range of experimental datasets we have used.

The heterogeneous reconstruction network consists of two encoders, an image encoder and a Gaussian encoder, and a single decoder that integrates their representations. The image encoder processes 2D particle images and transforms them into a continuous low-dimensional latent conformation space. Each point embedded into this low-dimensional manifold corresponds to a potential 3D conformation. Although the number of Gaussians is considerably smaller than that of voxels, directly inputting all Gaussian parameters into the network remains computationally infeasible. To address this challenge, GaussianEM introduces a learnable Gaussian encoding submodule to allow the network to capture the contextual relationships among Gaussians. The Gaussian encoder accepts attributes such as density, scale, and position for each Gaussian, and outputs a unique latent embedding that serves as its identifier. Inspired by NeRF-based architectures, positional encoding is incorporated to help the network capture high-frequency spatial information, thereby enhancing its ability to reconstruct fine structural details. The latent variable from the image encoder is then concatenated with the Gaussian embedding and passed to the decoder, which predicts the parameter variations for each Gaussian. This per-Gaussian paradigm allows efficient batch-level parallelization and facilitates the aggregation and discrimination of Gaussians according to their learned parameter changes.

In GaussianEM, the difference between the experimental particle images and the rendered projections (given known particle poses) is measured in the real-space domain. Although we still need to transform the projection twice to apply the CTF modulator, comparison in real space makes the resulting structural changes far more intuitive and interpretable. Once the network is trained, the learned conformation space can be visualized using standard dimensionality reduction techniques \cite[]{mcinnes2018umap}. By identifying cluster centers within this continuous latent space, representative 3D volumes can be generated to capture discrete conformational states. Moreover, since the conformation manifold is continuous, trajectories between clusters can be smoothly traversed to produce dynamic structural transitions. An additional advantage of the real-space Gaussian representation is its natural compatibility with atomic models. The displacements or loss of Gaussians are easily mapped to the atomic model through the nearest point connection. This property facilitates the generation of deformed atomic models (PDB structures) that correspond to different conformational states inferred from the cryo-EM data. Consequently, GaussianEM provides an optional perspective that links density-based heterogeneity analysis with atomic-level structures.

\subsection{Datasets and experimental details}
We applied GaussianEM to two simulation CryoBench datasets \cite[]{jeon2024cryobench} and five publicly available Electron Microscopy Public Image Archive (EMPIAR) \cite[]{iudin2016empiar} datasets. Four experimental datasets are presented in the main text, and an additional datasets are included in the Supplementary Material (sections ``$\alpha V\beta8$ integrin bound to latent TGF-$\beta$'').
For each dataset, consensus reconstruction and nonuniform refinement were performed in CryoSPARC \cite[]{punjani2017cryosparc} using all available experimental images. The estimated particle poses were fixed during training, and particle images were downsampled to a uniform size of $256\times256$. For datasets with smaller raw image sizes, the original resolution was retained. Initial Gaussian parameters were sampled from the consensus map after applying a contour-level threshold. When a corresponding PDB atomic model was available, Gaussian displacements were transferred to atomic coordinates to generate conformation-specific atomic models. The reconstructed density maps and atomic models were visualized in ChimeraX \cite[]{pettersen2021ucsf}. All experiments were conducted on an NVIDIA GeForce RTX 3080 with 10 GB RAM, with training typically completed within 3 to 10 hours, depending on the scale of the dataset. 

\begin{figure}[H]
	\centering
	\includegraphics[width=0.85 \textwidth]{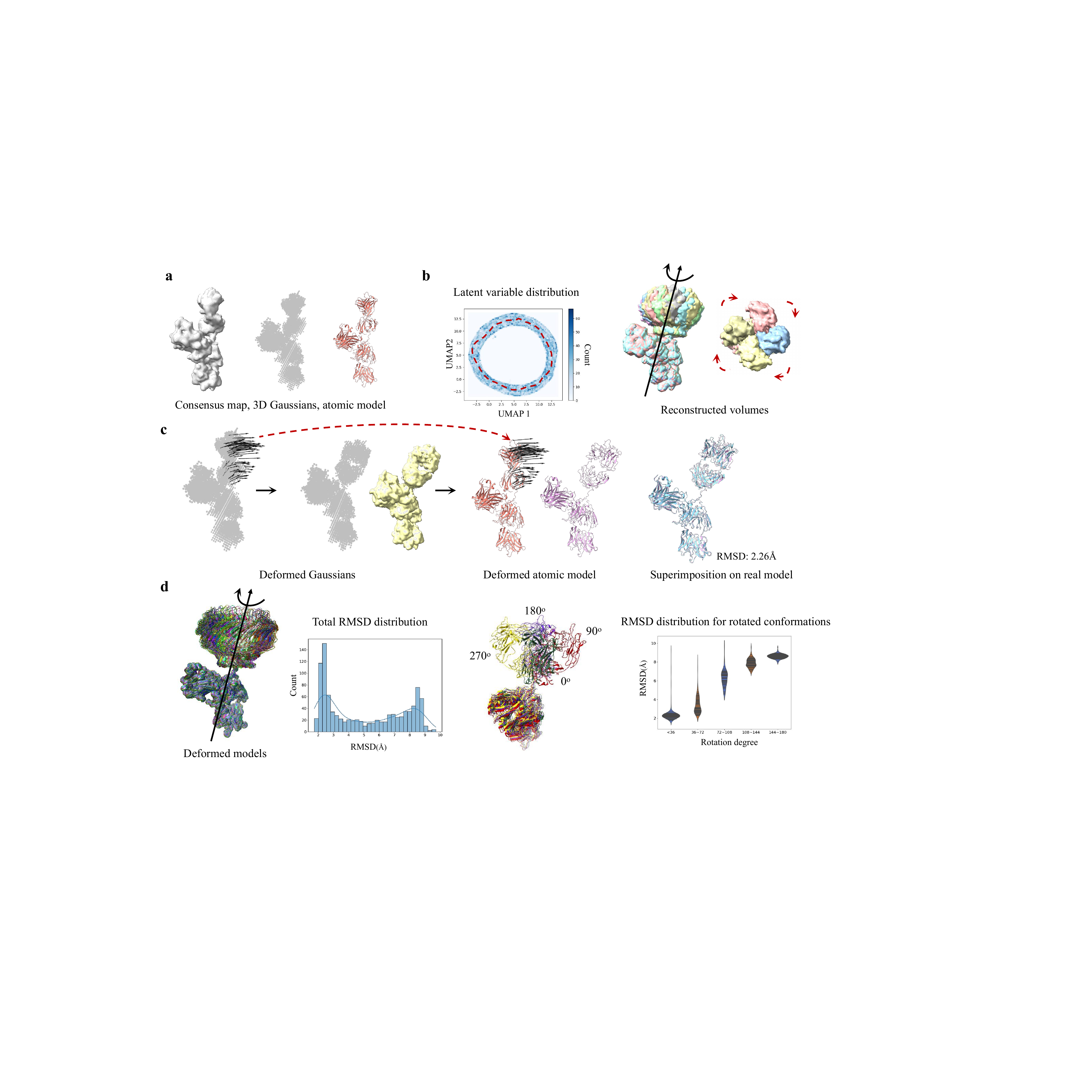}
	\caption{\textbf{GaussianEM captures large-scale motion in the simulated dataset.} \textbf{a} The consensus map and atomic model from the simulation dataset, along with the initial 3D Gaussians used in GaussianEM. \textbf{b} The two-dimensional latent conformation space after dimensionality reduction, exhibiting a circular distribution that corresponds to the continuous 3D structural motion. \textbf{c} The estimated Gaussian displacements are directly mapped onto atomic coordinates (red) in real space, generating the deformed atomic model (purple). The deformed model is superimposed on the provided ground-truth structure (blue), yielding an RMSD of 2.26 \AA, which demonstrates the accuracy of GaussianEM in capturing large-scale conformational motion. \textbf{d} Visualization of reconstructed atomic models. The left panel displays representative reconstructed atomic models. On the right, two RMSD distribution plots are shown: one for all estimated models and another categorized by rotation angles.}
	\label{fig:simulation}
\end{figure}

\subsection{Simulation data of large-scale motion}
The simulated IgG-1D dataset of the human immunoglobulin G (IgG) antibody complex from CryoBench was used to evaluate the ability of GaussianEM to capture large-scale molecular motions. This dataset models a one-dimensional continuous circular motion by rotating the dihedral angle that connects one of the fragment antibody (Fab) domains to the rest of the molecule. It includes 100 conformations, each with corresponding density maps and atomic models, and a total of 100,000 particle images with a box size of 128 pixels of width 3.0 \AA. Gaussian-based method e2gmm failed to describe this large-scale motion (see Figure \ref{fig:IgG} in Supplementary Material section ``IgG simulation'').
In contrast, GaussianEM successfully recovered the complete 360$^\circ$ rotational motion, as illustrated by the reconstructed volumes in Figure \ref{fig:simulation}b. The latent conformation space exhibits a distinct circular distribution, consistent with the continuous 3D rotational motion of the molecule. The motion is shown from two different viewing angles, including a top view aligned with the rotation axis. The reconstructed maps maintain comparable resolution across the full range of large-scale motions. Four representative conformations are displayed, each rotated by 90$^\circ$ along the rotational axis. 

We further mapped the estimated Gaussian displacements onto the atoms associated with the consensus structure, as illustrated by the red dashed lines in Figure \ref{fig:simulation}c. Unlike multi-body refinement, GaussianEM does not explicitly impose strict rigid-body constraints on the deformation fields, but the resulting atomic displacements still exhibit physically consistent motion. The Gaussian encoding paradigm, together with the local rigidity regularization term, collectively encourages neighboring Gaussians to undergo similar variations. While voxel-based methods can also capture rotational motions, they sometimes introduce inconsistencies in local regions (see Figure \ref{fig:IgG} in the Supplementary Material section ``IgG simulation”). 
In contrast, GaussianEM preserves the structural continuity of deformed volumes, as all deformations are applied to the same consensus model. The two superimposed atomic models show a high degree of structural overlap, with an RMSD of 2.26 \AA~ between corresponding atoms. This result indicates that GaussianEM can accurately capture large-scale conformational motion. Moreover, we reconstructed 1000 atomic models to measure the RMSD distribution, as shown in Figure \ref{fig:simulation}d. When categorized by rotation angles, the results show that conformations closer to the original structure yield smaller RMSD values, indicating higher reconstruction accuracy. For composite conformations within the 144–180$^{\circ}$ range, the average RMSD is approximately 8 \AA, which remains sufficient for capturing conformational heterogeneity.

To model more realistic structural heterogeneity, CryoBench also introduces the IgG-RL dataset, which is generated by modeling the 5-residue linker through backbone dihedral angle sampling based on disordered peptide Ramachandran distributions. This dataset comprises 100 conformations, with 1,000 projections generated for each conformation at a box size of 128 pixels and pixel size of 3.0 \AA. Voxel-based CryoDRGN displayed occasional loss of density in structural domains of the IgG-RL dataset exhibiting disordered dynamics (see Figure \ref{fig:IgG_RL}c in the Supplementary Material section ``IgG simulation”). In contrast, GaussianEM can recover conformational motions exhibiting stochastic variations (see Figure \ref{fig:IgG_RL}b in the Supplementary Material, section ``IgG simulation'').
\subsection{Ribosome assembly}

\begin{figure}[H]
	\centering
	\includegraphics[width=0.85 \textwidth]{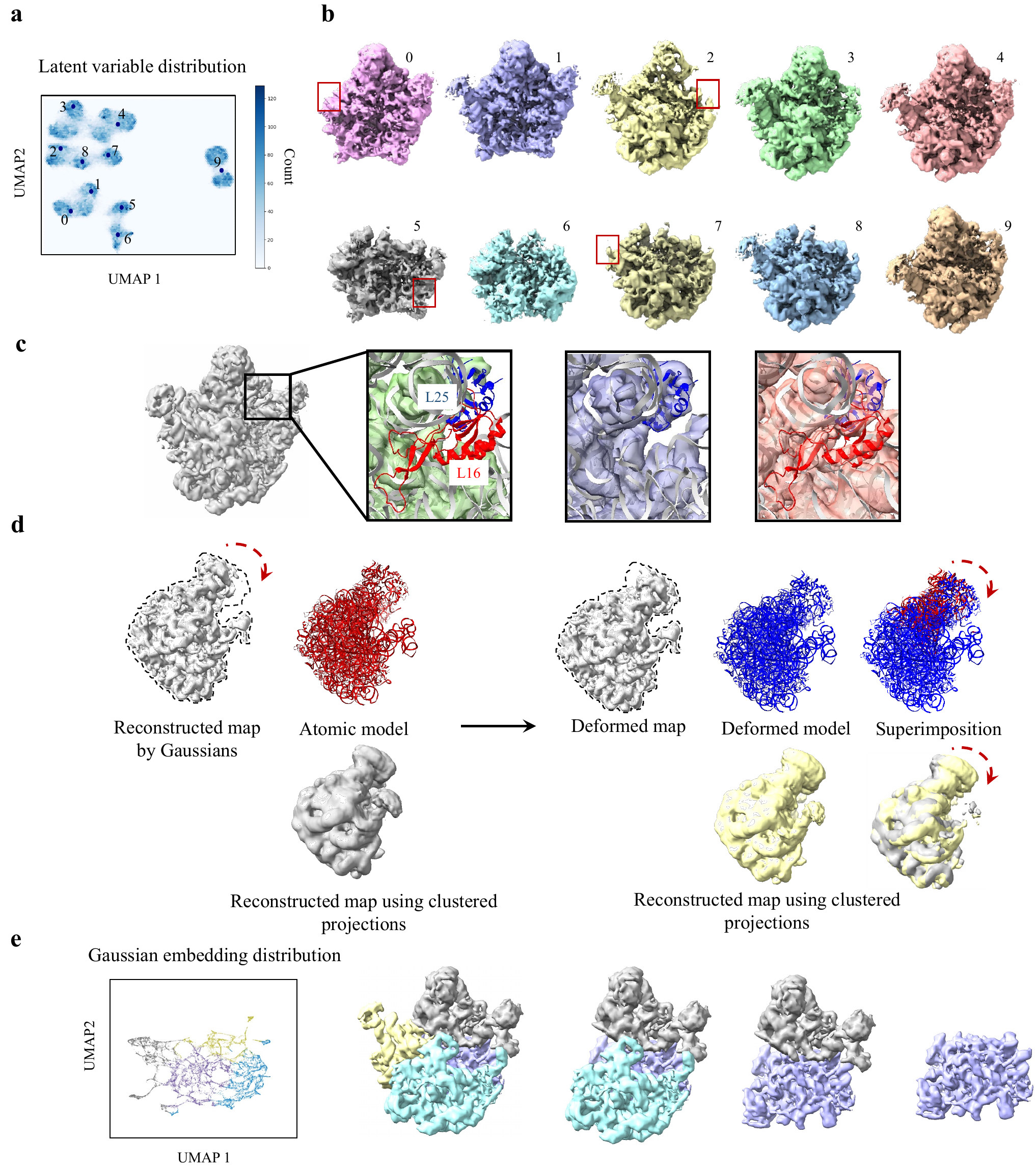}
	\caption{\textbf{Heterogeneity analysis of Ribosome assembly.} \textbf{a} Visualization of the latent space encoded by GaussianEM, where dark blue dots denote the cluster centers corresponding to distinct conformational classes. \textbf{b} Ten representative density maps generated by GaussianEM from the cluster centers, each labeled with its class index. Regions showing major conformational differences are highlighted in red boxes. \textbf{c} Examples of ribosomal protein L16, L25 that are present, partially present and fully missing. \textbf{d} Visualization of the motion of the dynamic central protuberance domain alongside the corresponding changes in the atomic models and reconstructed map by CryoSPARC. A dotted envelope is fixed to better illustrate structural motion. \textbf{e} The distribution of learned Gaussian embeddings. Four distinct clusters are identified and reconstructed. On the right, the corresponding volumes are shown, with each volume illustrating the effect of removing one component.}
	\label{fig:10841}
\end{figure}

The bL17-limited E. coli ribosome assembly intermediates (EMPIAR-10841) \cite[]{rabuck2022quantitative} contains a large number of discrete conformations, which have been validated by both classification-based and deep learning–based analyses \cite[]{zhong2021cryodrgn, rabuck2022quantitative}. The dataset consists of 123804 particle images with a box size of 160 pixels of width 2.64 \AA. In our experiment, we first allowed both the density and position parameters of the Gaussians to vary, without applying any masking operation. Uniform Manifold Approximation and Projection (UMAP) \cite[]{mcinnes2018umap} was used to reduce the latent variables to the 2D space to visualize the conformation distributions. The resulting latent space shows a clear separation of particle clusters, from which we select ten representative cluster centers for detailed analysis. As shown in Figure \ref{fig:10841}b, distinct structural differences are observed among the clusters, while subtle compositional changes are highlighted by the red boxes. The large-scale structural differences likely involve the ribosomal base and several stalk regions, reflecting different stages or pathways of ribosome assembly \cite[]{rabuck2022quantitative}. The presence of smaller local regions may correspond to the incorporation of ribosomal RNA (rRNA) segments during the maturation process. Class 9 appears distant from other clusters in the 2D latent space, and its reconstructed volume exhibits a different orientation, likely due to minor errors in pose estimation. We further investigated subtle structural variability within the dataset, as illustrated in Figure \ref{fig:10841}c. The example highlights ribosomal proteins L16 and L25, which appear in varying states—fully present, partially present, or completely absent—across different conformations.

Following the smearing effect observed in the dynamic central protuberance (CP) domain \cite[]{chen2021deep}, we subsequently constrained the variation to position parameters within that region. A clear tilting motion of that domain is revealed, as shown in Figure \ref{fig:10841}d. For better visualization, we aligned the PDB 6PJ6 atomic model to the reconstructed volume and mapped the Gaussian displacements onto the corresponding atoms. The superimposed atomic models are displayed alongside the reconstructed volumes, illustrating the detailed conformational changes captured by GaussianEM. We further analyzed the learned Gaussian embeddings by performing k-means clustering. Four clusters were identified and colored in Figure \ref{fig:10841}e. By removing each cluster in turn, we observed that the remaining structure closely resembled the corresponding discrete conformations, indicating that the Gaussian embeddings effectively capture different structural features.

\subsection{Pre-catalytic spliceosome}

\begin{figure}[H]
	\centering
	\includegraphics[width=0.85 \textwidth]{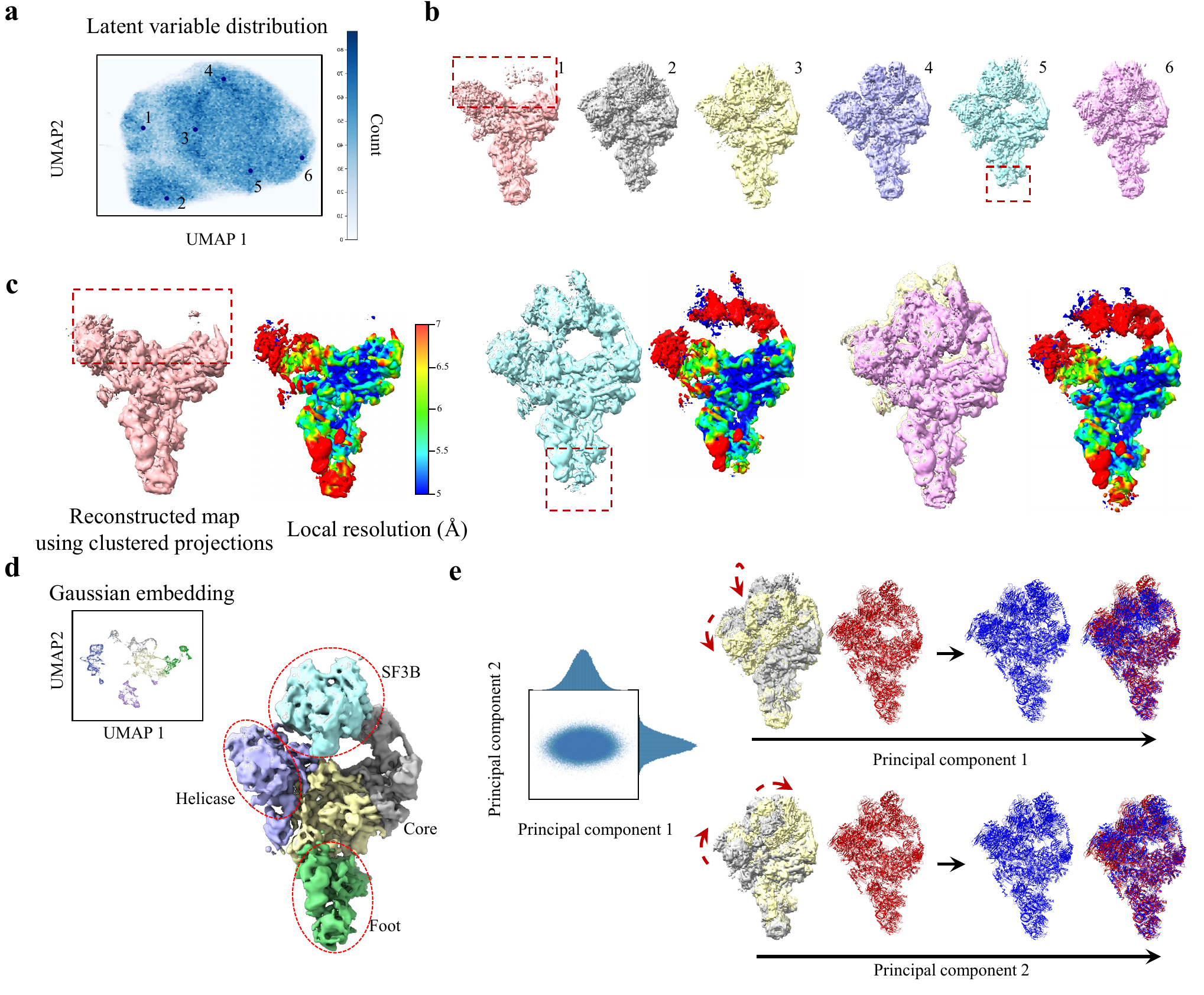}
	\caption{\textbf{Heterogeneity analysis of precatalytic spliceosome.} \textbf{a} Visualization of the latent space encoded by GaussianEM \textbf{b} Six representative density maps generated by GaussianEM from the cluster centers, each labeled with its class index. Major conformational differences are highlighted in red boxes. \textbf{c} Reconstructed maps by CryoSPARC using clustered projection from GaussianEM. \textbf{d} Visualization of segmented Gaussians. Different colored components roughly respond to different biological domains. \textbf{e} UMAP visualization of principal components 1, 2 and 3. The reconstructed volumes and their corresponding atomic models are shown on the right.}
	\label{fig:10180}
\end{figure}

GaussianEM was further applied to analyze structural variations in the pre-catalytic spliceosome dataset (EMPIAR-10180) \cite[]{plaschka2017structure}. This dataset has 327490 particle images with a box size of 320 pixels of width 1.699 \AA. The images were downsampled to $256\times256$, and initial Gaussians were sampled from the consensus map deposited in EMPIAR-10180. All Gaussian parameters were allowed to vary during training. Using the K-means algorithm, six distinct clusters were identified (Figure \ref{fig:10180}a). Reconstruction of density maps from the cluster centers revealed clear compositional differences and continuous motion trajectories. Notably, the absence of the SF3b and foot domains is highlighted by the red boxes in Figure \ref{fig:10180}b. Similar compositional heterogeneity was captured by the NeRF-based approach OPUS-DSD \cite[]{luo2023opus} but not by the Gaussian-based method e2gmm \cite[]{chen2021deep}. Clustered projections were further reconstructed in CryoSPARC to validate the effectiveness of GaussianEM (Figure~\ref{fig:10180}c). The absence of SF3B and foot domains is also clearly visible.
The learned Gaussian embeddings were grouped into five clusters, as shown in Figure \ref{fig:10180}d. Each cluster roughly corresponds to a distinct biological domain. The blue cluster predominantly represents the SF3b region, while the purple cluster mainly corresponds to the helicase domain.
Furthermore, PCA of the latent space reveals distinct patterns of conformational variability along the first two principal components (PC1–PC2) in Figure \ref{fig:10180}e. The helicase and SF3b domains exhibit coordinated but distinct rotational motions along different components. Along the first principal component, the spliceosome transitions from an “open” to a “closed” state: at the beginning of the trajectory, the SF3b domain stays on the top of the core body, and at the end, it folds downward toward the core region. Along the second principal component, the helicase and SF3b domains jointly display a left–right swinging motion. The displacements of Gaussians were mapped to the atomic model PDB-5NRL, which was fitted to density maps by CryoAlign \cite[]{he2024accurate}. The resulting atomic motions were smooth and coherent, indicating that our method can suppress artificial structural distortions to some extent. Voxel-based methods can produce overall continuous trajectories but may exhibit local inconsistencies (see Figure \ref{fig:10180_drgn} in Supplementary Material section ``Pre-catalytic spliceosome'').

\subsection{T6SS effectors}

\begin{figure}[H]
	\centering
	\includegraphics[width=0.85 \textwidth]{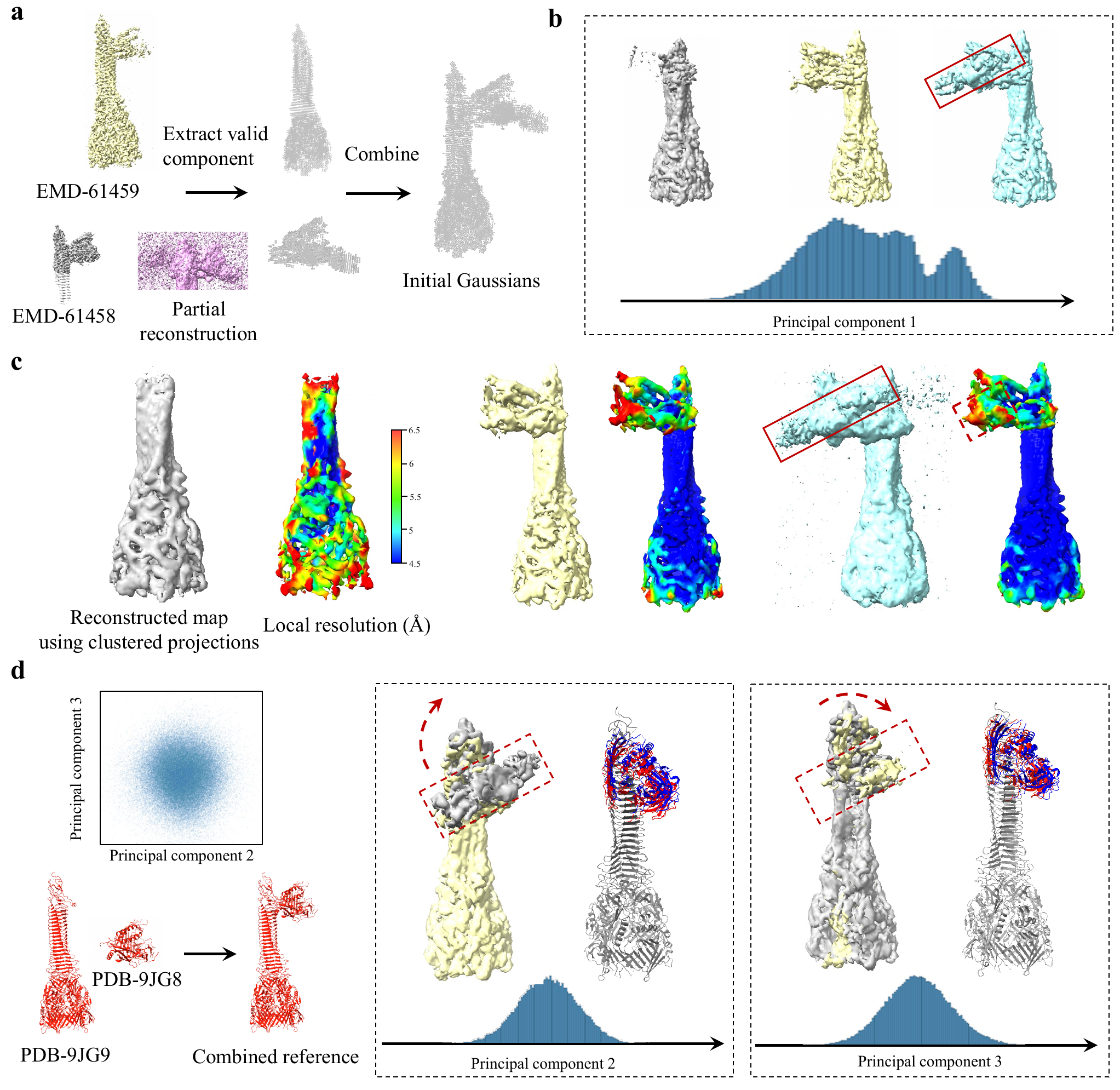}
	\caption{\textbf{Heterogeneity analysis of BtpeA-BtaeB-VgrG complex.} \textbf{a} Generation of the initial Gaussians. Valid points were extracted from two density maps (EMD-61459 and EMD-61458) and combined to form the initial model. \textbf{b} Three reconstructed volumes along the first principal component. A region not present in publicly available data is highlighted by a red box. \textbf{c} Reconstructed maps by CryoSPARC using clustered projection from GaussianEM. \textbf{d} UMAP visualization of principal components 2 and 3. The reconstructed volumes and their corresponding atomic models are shown on the right.}
	\label{fig:gao}
\end{figure}

We then used the cargo complex (BtpeA-BtaeB-BtapC) bound to the VgrG spike \cite[]{li2025conserved} to test GaussianEM.
This dataset consists of 106565 particle images with a box size of 480 pixels of width 1.18 \AA. In contrast to the previous datasets, this dataset lacked a complete density for Gaussian initialization. Two related maps, EMD-61458 and EMD-61459, each represented partial structures. The homogeneous reconstruction from CryoSPARC also provided additional structural information. We extracted valid points from these maps and aligned them using CryoAlign to generate a unified initial Gaussian model. The particle images were downsampled to $256\times256$, and the density and position parameters of Gaussians were permitted to change. 

After training, we performed PCA on the latent space to explore conformational variability. The first principal component revealed discrete conformational transitions, as illustrated in Figure \ref{fig:gao}b. Three representative volumes were reconstructed, two of which correspond to known structures. The gray volume represents the main body of the VgrG spike, consistent with the atomic model PDB-9JG9. The yellow conformation includes the BtpeA and BtaeB effectors assembled on the base of the gray structure, corresponding to PDB-9JG8. The blue conformation further extends the state, maintaining the additional structural region mentioned above and highlighted by the red box. To validate these results, the three clustered projections were reconstructed in CryoSPARC (Figure \ref{fig:gao}c). Although the additional region remains partially unresolved, the reconstructions reveal clearer envelope features around the effectors, as indicated by the dashed red box.
Moreover, the second and third principal components in Figure \ref{fig:gao}d capture continuous motions within the effector region. This change was easily neglected (see Figure \ref{fig:gao_drgn} in Supplementary Material section ``T6SS effectors''). The reconstructed volumes along these components reveal smooth structural transitions rather than discrete states. Gaussian displacements were mapped onto the assembled atomic models, where stable regions are colored gray and flexible regions are shown in red and blue for visualization. Interestingly, the reconstructed volumes along the second component differ from those along the third: the former retain additional peripheral structures, while the latter preserve the core architecture of the two effectors and the VgrG spike. 

\subsection{$\alpha V\beta8$ integrin}

\begin{figure}[H]
	\centering
	\includegraphics[width=0.85 \textwidth]{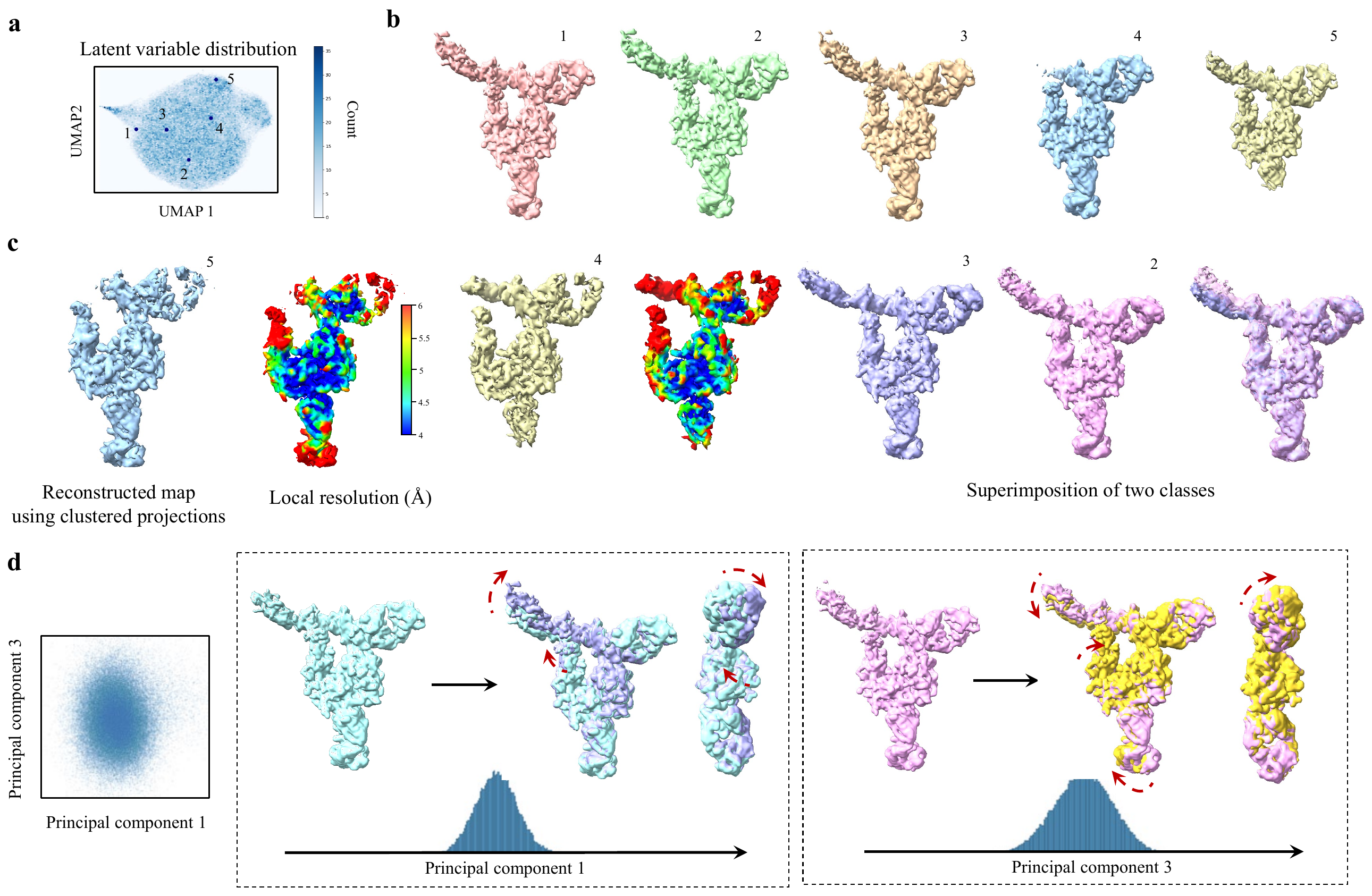}
	\caption{\textbf{Heterogeneity analysis of $\alpha V\beta8$ integrin bound to two Fabs.} \textbf{a} Visualization of the latent space encoded by GaussianEM \textbf{b} Five representative density maps generated by GaussianEM from the cluster centers, each labeled with its class index. \textbf{c} Reconstructed maps by CryoSPARC using clustered projection from GaussianEM. \textbf{d} UMAP visualization of principal components 1 and 3. The reconstructed volumes and their superimposition are shown on the right.}
	\label{fig:10345}
\end{figure}

The $\alpha V\beta8$ integrin is a highly flexible protein involved in cell differentiation during development (EMPIAR-10345) \cite[]{campbell2020cryo}. This dataset consists of 84266 particle images with a box size of 300 pixels of width 1.35 \AA. The flexible arm of the integrin is difficult to resolve through conventional homogeneous reconstruction; therefore, we used the density map refined by 3DFlex \cite[]{punjani20233dflex} as our initial model. All Gaussian parameters were allowed to vary during training. Using K-means clustering, five representative volumes were identified and reconstructed in Figure \ref{fig:10345}b. In addition to the arm-swing motion previously reported \cite[]{punjani20233dflex}, GaussianEM also revealed the absence of the arm and the foot of the core protein. These findings were validated by homogeneous reconstruction in CryoSPARC using clustered projections (Figure \ref{fig:10345}c). Furthermore, with the help of compact clustering, the motion of the arm is also visible in CryoSPARC reconstruction. Figure \ref{fig:10345} illustrates the detailed conformational changes of $\alpha V\beta8$ integrin along the first and third principal components.


\subsection{Quantitative comparisons}
Quantifying the performance of heterogeneity analysis methods remains a major challenge in computational structural biology. Homogeneous structure determination benefits from established benchmarks such as the Fourier Shell Correlation (FSC) curve. In contrast, heterogeneous reconstruction lacks universally accepted evaluation metrics due to the inherent continuity and complexity of the conformational space. To address this limitation, several quantitative measures have been proposed, ranging from conformational diversity to voxel-level consistency. Building on these metrics, we performed a comprehensive quantitative comparison of GaussianEM with representative heterogeneous reconstruction methods, including CryoDRGN \cite[]{zhong2021cryodrgn}, DynaMight \cite[]{schwab2023dynamight}, and 3DFlex \cite[]{punjani20233dflex}.
\subsubsection{Completeness of heterogeneity analysis}
The Flatiron Institute (FI) Cryo-EM Conformational Heterogeneity Challenge \cite[]{astore2025inaugural} attracted broad participation from leading research groups and introduced a novel set of evaluation metrics for conformational heterogeneity. We adopt one of its key metrics, Percentage of Captured Variance (PCV), to quantitatively assess the ability of different methods to characterize conformational motions. PCV is a pairwise comparison metric that measures the overlap between two extracted latent subspaces. Specifically, when the conformational set produced by method A is treated as the reference, and that of method B as the test set, the PCV value PCV(A, B) is defined as:
\begin{align}
\text{PCV}(A, B) = \frac{\|\mathbf{V}_B^T\mathbf{V}_A\mathbf{\mathbf{S}}_A\|_F^2}{\|\mathbf{V}_A^T\mathbf{V}_A\mathbf{\mathbf{S}}_A\|_F^2} = \frac{\sum_{l,k}[\mathbf{S}_A]_k^2([\mathbf{V}_B]_l^T[\mathbf{V}_A]_k)^2}{\sum_k[\mathbf{S}_A]_k^2}
\end{align}
where $\mathbf{V}_A$ and $\mathbf{S}_A$ denote the eigenvector matrix and the diagonal eigenvalue matrix of the reference set’s covariance matrix, respectively, $\mathbf{V}_B$ represents the eigenvector matrix of the test set’s covariance matrix, and $\|\cdot\|_F$ is Frobenius norm. The PCV metric measures how much of the weighted conformational subspace of a reference set can be projected onto the subspace spanned by a test set. The resulting ratio, which ranges from 0 to 1, represents the fraction of conformational variance in the reference set that is captured by the test set. A key property of PCV is its asymmetry, which enables comparative evaluation between methods. Method B is considered superior to Method A when $\text{PCV}(A, B)$ exceeds $\text{PCV}(B, A)$. This indicates that the conformational space of Method B captures most of the variance present in Method A, while Method A fails to explore the conformations unique to Method B. Such a relationship indirectly demonstrates that Method B provides a more comprehensive and diverse characterization of the conformational landscape.

\begin{table}[!ht]
	\centering
	\setlength{\tabcolsep}{1.4mm}{
            \renewcommand\arraystretch{1.}
		\begin{tabular}{ccccccccc}	
                \hline
			  \toprule
            \toprule
            \multicolumn{1}{c}{\multirow{2}{*}{Method}}&\multicolumn{4}{c}{IgG-1D}&\multicolumn{4}{c}{IgG-RL}\\
            \cmidrule(lr){2-5}\cmidrule(lr){6-9}
            \multicolumn{1}{c}{~}& CryoDRGN&DynaMight&3DFlex&GaussianEM& CryoDRGN&DynaMight&3DFlex&GaussianEM\\
            \hline
                \multicolumn{1}{c}{GT}& 0.8091& --& 0.8177&0.8258&0.7009&--& 0.6294&0.7320\\ 
            \bottomrule
            \bottomrule
		\end{tabular}
	}
    \caption{PCV result on IgG datasets. "--" indicates that the method fails to produce a valid result.}
    \label{tab:pcv_IgG-1D}
    \vspace{-1.em}
\end{table}

We first evaluated the performance of compared methods on the simulation IgG-1D and IgG-RL datasets, for which ground-truth (GT) conformations are available. For each method, an ensemble of 100 conformations was generated by applying K-means clustering to the latent vectors and decoding the cluster centroids into 3D volumes. As summarized in Table \ref{tab:pcv_IgG-1D}, GaussianEM yields a more diverse conformational ensemble than CryoDRGN and 3DFlex. In contrast, DynaMight fails to effectively resolve conformations in datasets with large-scale structural motions.
\begin{table*}[!ht]
\footnotesize
		\begin{tabular}{cp{1.4cm}<{\centering}p{1.4cm}<{\centering}p{1.4cm}<{\centering}p{1.4cm}<{\centering}p{0.01cm}<{\centering}p{1.4cm}<{\centering}p{1.4cm}<{\centering}p{1.4cm}<{\centering}p{1.4cm}<{\centering}}	
        \hline
            \toprule
            \toprule
            \multicolumn{1}{c}{\multirow{2}{*}{Method}}&\multicolumn{4}{c}{EMPIAR-10180}&\multicolumn{1}{c}{~}&\multicolumn{4}{c}{EMPIAR-10343}\\
            \cmidrule(lr){2-5} \cmidrule(lr){7-10} 
            \multicolumn{1}{c}{~}&CryoDRGN&DynaMight&3DFlex&GaussianEM& \multicolumn{1}{c}{~}&CryoDRGN&DynaMight&3DFlex&GaussianEM \\
            \cmidrule(lr){1-5} \cmidrule(lr){7-10}
            CryoDRGN & 1& \cellcolor{group3}0.4142& \cellcolor{group2}0.3306& \cellcolor{group1dark}0.4569& \multicolumn{1}{c}{~}& 1& \cellcolor{group3}0.0875& \cellcolor{group2}0.1225& \cellcolor{group1dark}0.1425\\
            DynaMight & \cellcolor{group3}0.5686& 1& \cellcolor{group1}0.5206& \cellcolor{group2dark}0.5245& \multicolumn{1}{c}{~}& \cellcolor{group3}0.1214& 1& \cellcolor{group1}0.2404& \cellcolor{group2dark}0.1998\\
            3DFlex & \cellcolor{group2}0.5619& \cellcolor{group1}0.6218& 1& \cellcolor{group3dark}0.5574& \multicolumn{1}{c}{~}& \cellcolor{group2}0.3054& \cellcolor{group1}0.4066& 1& \cellcolor{group3dark}0.5227\\
            GaussianEM & \cellcolor{group1dark}0.3880& \cellcolor{group2dark}0.2908& \cellcolor{group3dark}0.2489& 1& \multicolumn{1}{c}{~}& \cellcolor{group1dark}0.1364& \cellcolor{group2dark}0.1684& \cellcolor{group3dark}0.2499& 1\\ \hline
            \multicolumn{1}{c}{\multirow{2}{*}{Method}}&\multicolumn{4}{c}{EMPIAR-10345}&\multicolumn{1}{c}{~}&\multicolumn{4}{c}{T6SS effectors} \\ 
            \cmidrule(lr){2-5} \cmidrule(lr){7-10}
            \multicolumn{1}{c}{~}&CryoDRGN&DynaMight&3DFlex&GaussianEM& \multicolumn{1}{c}{~}&CryoDRGN&DynaMight&3DFlex&GaussianEM \\ 
            
            \cmidrule(lr){1-5} \cmidrule(lr){7-10}
             CryoDRGN & 1& \cellcolor{group3}0.2045& \cellcolor{group2}0.2365& \cellcolor{group1dark}0.2495& \multicolumn{1}{c}{~}& 1& \cellcolor{group3}0.0703& \cellcolor{group2}0.0607& \cellcolor{group1dark}0.1193\\
            DynaMight & \cellcolor{group3}0.2266& 1& \cellcolor{group1}0.4851& \cellcolor{group2dark}0.3129&  \multicolumn{1}{c}{~}&\cellcolor{group3}0.0808& 1& \cellcolor{group1}0.1705& \cellcolor{group2dark}0.1081\\
            3DFlex & \cellcolor{group2}0.4156& \cellcolor{group1}0.6609& 1& \cellcolor{group3dark}0.3833& \multicolumn{1}{c}{~}& \cellcolor{group2}0.0328& \cellcolor{group1}0.2266& 1& \cellcolor{group3dark}0.0939\\
            GaussianEM & \cellcolor{group1dark}0.2054& \cellcolor{group2dark}0.2281& \cellcolor{group3dark}0.1732& 1& \multicolumn{1}{c}{~}& \cellcolor{group1dark}0.0536& \cellcolor{group2dark}0.0685& \cellcolor{group3dark}0.0772& 1\\ 
            \bottomrule
            \bottomrule
		\end{tabular}
 \caption{PCV results on EMPIAR datasets. The colors correspond to the following comparisons: light green for CryoDRGN vs. DynaMight, light orange for CryoDRGN vs. 3DFlex, light blue for DynaMight vs. 3DFlex, dark green for 3DFlex vs. GaussianEM, dark orange for DynaMight vs. GaussianEM, and dark blue for CryoDRGN vs. GaussianEM.}
\label{tab:pcv2}
\end{table*}

Next, we extended our evaluation to the experimental EMPIAR datasets described in the main text. For each method, the conformational space was represented by 50 reconstructed conformations. As shown in Table \ref{tab:pcv2}, the comparison between GaussianEM and CryoDRGN (dark blue) indicates that GaussianEM recovers a broader and more diverse conformational space. For example, in the Supplementary Material section ``T6SS effectors'', GaussianEM successfully captures the slight conformational motions of effectors that are not resolved by CryoDRGN. Similarly, the comparisons highlighted by light green and light blue demonstrate the superiority of CryoDRGN over DynaMight and of DynaMight over 3DFlex, respectively. Overall, these results establish a consistent performance ranking: GaussianEM $>$ CryoDRGN $>$ DynaMight $>$ 3DFlex. A similar ranking was observed across the other EMPIAR datasets. 
\begin{table*}[!ht]
\footnotesize	
		\begin{tabular}{cp{1.4cm}<{\centering}p{1.4cm}<{\centering}p{1.4cm}<{\centering}p{1.4cm}<{\centering}p{0.01cm}<{\centering}p{1.4cm}<{\centering}p{1.4cm}<{\centering}p{1.4cm}<{\centering}p{1.4cm}<{\centering}}	
        \hline
            \toprule
            \toprule
            \multicolumn{1}{c}{\multirow{2}{*}{Method}}&\multicolumn{4}{c}{FI First  Cryo-EM dataset}&\multicolumn{1}{c}{~}&\multicolumn{4}{c}{FI Second Cryo-EM dataset}\\
            \cmidrule(lr){2-5} \cmidrule(lr){7-10}
            \multicolumn{1}{c}{~}&CryoDRGN&DynaMight&3DFlex&GaussianEM& \multicolumn{1}{c}{~}&CryoDRGN&DynaMight&3DFlex&GaussianEM \\
            \cmidrule(lr){1-5} \cmidrule(lr){7-10}
            CryoDRGN & 1& \cellcolor{group3}--& \cellcolor{group2}0.1819& \cellcolor{group1dark}0.0824 & \multicolumn{1}{c}{~}& 1& \cellcolor{group3}--& \cellcolor{group2}0.0163& \cellcolor{group1dark}0.1300\\
            DynaMight & \cellcolor{group3}--& 1& \cellcolor{group1}--& \cellcolor{group2dark}-- & \multicolumn{1}{c}{~}& \cellcolor{group3}--& 1& \cellcolor{group1}--& \cellcolor{group2dark}--\\
            3DFlex & \cellcolor{group2}0.2868& \cellcolor{group1}--& 1& \cellcolor{group3dark}0.2376& \multicolumn{1}{c}{~}&\cellcolor{group2}0.0231& \cellcolor{group1}--& 1& \cellcolor{group3dark}0.0148\\
            GaussianEM & \cellcolor{group1dark}0.0396& \cellcolor{group2dark}--& \cellcolor{group3dark}0.1132& 1& \multicolumn{1}{c}{~}&\cellcolor{group1dark}0.0363& \cellcolor{group2dark}--& \cellcolor{group3dark}0.0050&1\\ 
            \bottomrule
            \bottomrule
		\end{tabular}
 \caption{PCV results on FI datasets. The colors correspond to the following comparisons: light green for CryoDRGN vs. DynaMight, light orange for CryoDRGN vs. 3DFlex, light blue for DynaMight vs. 3DFlex, dark green for 3DFlex vs. GaussianEM, dark orange for DynaMight vs. GaussianEM, and dark blue for CryoDRGN vs. GaussianEM. And color scheme is consistent with Table \ref{tab:pcv2}. "--" indicates that the method fails to produce a valid result.}
\label{tab:pcv1}
\end{table*}

Finally, to further assess robustness and generalizability, we performed PCV comparisons on two datasets from the FI cryo-EM heterogeneity challenge, both of which exhibit complex conformational heterogeneity of thyroglobulin (Table~\ref{tab:pcv1}). The first dataset is experimental and contains 33,742 particle images, while the second is a simulated dataset of the same size, generated by projecting 3,861 distinct conformations derived from molecular dynamics simulations. The pairwise PCV results consistently show that GaussianEM outperforms CryoDRGN in resolving conformational heterogeneity, with CryoDRGN in turn outperforming 3DFlex. Notably, DynaMight failed to produce plausible conformations on either dataset under default parameters.
 
\subsubsection{FSC-based metrics on volume generation}
\begin{figure}[H]
	\centering
	\includegraphics[width= 0.75\textwidth]{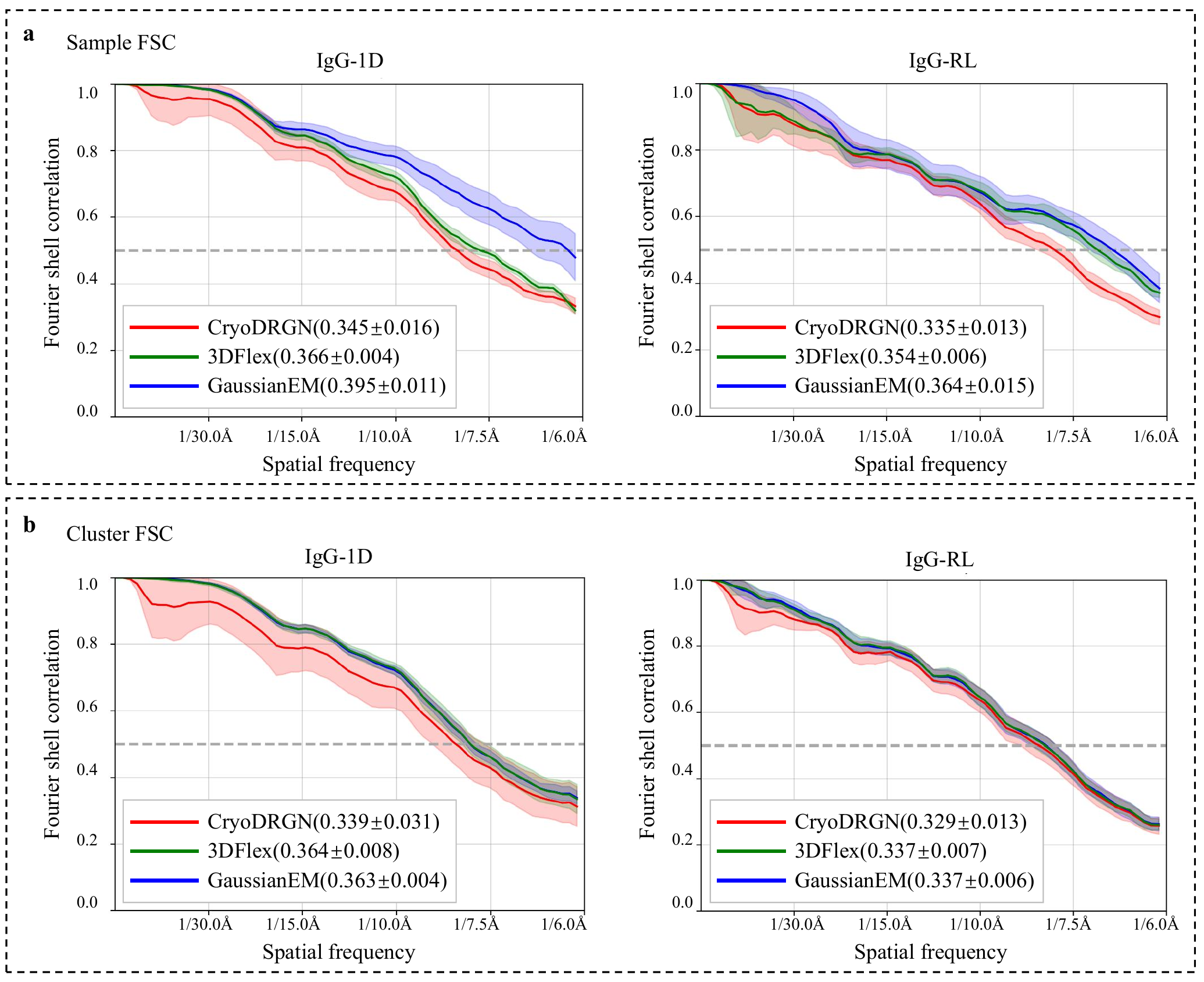}
	\caption{\textbf{a} Sample-FSC results of compared methods on IgG datasets. \textbf{b} Cluster-FSC results of compared methods on IgG dataset.}
	\label{fig:IgG_FSC}
\end{figure}

CryoBench introduced an FSC-based metric, termed Sample FSC, to evaluate the reconstructed volumes. Sample FSC is computed by first applying K-means clustering to latent embeddings. The cluster centroids are then decoded into representative volumes. Because these unsupervised reconstructions do not directly correspond to GT volumes, the Area Under the Fourier Shell Correlation curve ($FSC_{AUC}$) is computed between each reconstructed volume and all GT volumes. The maximum value is retained as the score for each reconstruction. The distribution of these maximum scores across clusters reflects the overall generative performance of the model. Based on this idea, we introduce a new metric, termed Cluster FSC. Unlike Sample FSC, Cluster FSC clusters 2D projection images using their corresponding latent embeddings and subsequently reconstructs representative 3D volumes for each cluster by back-projection. While Sample FSC primarily evaluates the generative quality of the decoder, Cluster FSC provides a more rigorous assessment of the model's classification and disentanglement capabilities. By reconstructing volumes directly from clustered projection images rather than from decoded latent variables, Cluster FSC metric reduces the impact of potential decoder-induced artifacts or biases. This design provides a more reliable assessment of whether conformational heterogeneity is correctly disentangled.

In this section, we use $K = 20$ for all clustering tasks. Figure \ref{fig:IgG_FSC} shows the Sample FSC and Cluster FSC results for GaussianEM, 3DFlex, and CryoDRGN on the IgG-1D and IgG-RL datasets. As shown in Figure \ref{fig:IgG_FSC}a, GaussianEM achieves higher reconstruction quality than CryoDRGN and 3DFlex, which is consistent with the visual results of decoded volumes (see Figure \ref{fig:IgG}d and Figure \ref{fig:IgG_RL}c in the Supplementary Material section ``IgG simulation”).
Figure \ref{fig:IgG_FSC}b illustrates that GaussianEM outperforms CryoDRGN in conformational disentanglement and achieves performance comparable to 3DFlex. These results are consistent with the latent variable distributions exhibited in Figure \ref{fig:simulation}b, Figure \ref{fig:IgG}c and Figure \ref{fig:IgG_RL} (presented in the Supplementary Material section ``IgG simulation”).

\section{Discussion}

In this study, we introduce GaussianEM, a Gaussian-based framework designed to intuitively capture both compositional and conformational heterogeneity in cryo-EM datasets. Most recent cryo-EM heterogeneity analysis methods typically rely on voxel-based representations of macromolecular structures. Such representations allow for the identification of compositional variations by comparing density between reconstructed volumes. However, they are often limited in describing continuous conformational motion. Tracking subtle structural changes using voxel grids usually requires manual interpretation and lacks smoothness. Additionally, this process sometimes introduces artifacts. In contrast, Gaussian-based models offer a more natural way to describe structural variability. In this formulation, changes in Gaussian density directly correspond to the presence or absence of components, while Gaussian displacements provide an intuitive visualization of continuous motion. This representation makes it easier to locate and interpret regions undergoing change. Moreover, the isotropic Gaussian function ensures consistent behavior in both two and three dimensions, linking 2D projections directly to 3D structural deformations. These properties form the foundation of GaussianEM in exploring molecular heterogeneity.

Despite the advantages of Gaussian representation, designing and training an effective network architecture remains challenging. The first question lies in how to enable the network to recognize and process 3D Gaussians. Most existing approaches employ a variational autoencoder framework to bridge the 2D–3D mapping, a strategy that GaussianEM also adopts. However, embedding Gaussians into such a framework is nontrivial. For instance, e2gmm leverages Fourier ring correlation to derive Gaussian gradients and concatenates them as network inputs, predicting parameter variations for all Gaussians. This design allows the network to learn from experimental images, but it demands extensive GPU memory, severely limiting the number of Gaussian spheres and thus the achievable resolution. DynaMight, on the other hand, follows the NeRF paradigm and incorporates positional encoding to predict Gaussian displacements. Its formulation is primarily suited for modeling continuous motions, where positional information alone is often sufficient. To overcome these limitations and improve scalability, GaussianEM introduces an additional Gaussian encoder module. This encoder accepts the intrinsic parameters of each Gaussian—such as position, scale, and density—and outputs a unique latent embedding that serves as its identifier. This design allows the number of Gaussians to scale independently of the overall network architecture, with GPU usage increasing only linearly. In practice, GaussianEM can accommodate tens of thousands of Gaussians, enabling a fine-grained representation. At the same order of magnitude, variations in Gaussian parameters can be naturally translated into atomic conformations.

The second question is controlling the variations of 3D Gaussians. In GaussianEM, the loss is computed between the experimental images and the Gaussian projections in real space. Due to the presence of noise and contrast variation in experimental cryo-EM images, it is necessary to regulate Gaussian variations to prevent overfitting. Molecular heterogeneity often exhibits local coherence, meaning that spatially adjacent regions tend to undergo correlated structural changes. To capture this property, GaussianEM introduces constraints on the latent variables of Gaussians, encouraging neighboring Gaussians to deform in a consistent manner. When modeling continuous conformational motions, we further enforce local rigidity to ensure smooth and physically plausible displacements. Without such constraints, we observed that unconstrained Gaussian displacements tend to cluster together to fit density values, rather than adjusting their individual densities appropriately.

One limitation is that GaussianEM needs a reasonable consensus map as the initial model. The consensus model must include all regions of interest. This consensus reconstruction must include all regions of interest; otherwise, regions lacking sufficient density information cannot be represented by corresponding Gaussian spheres. In practice, missing regions can sometimes be recovered through mask-based operations or local refinements. Then, several maps can be combined to acquire a more 'complete' initial model. However, such composite models may introduce pose or shape inconsistencies among particles, making it challenging for GaussianEM to accurately interpret and learn structural variations.

In conclusion, GaussianEM offers an intuitive and interpretable framework for analyzing both compositional and conformational heterogeneity in cryo-EM datasets. By modeling macromolecular structures with 3D Gaussians, it bridges the gap between density-based analysis and atomic-level interpretation. When integrated with corresponding atomic models, GaussianEM can generate continuous atomic conformational landscapes. This approach allows researchers to gain deeper insights into the dynamic structural behavior of macromolecules and their functional mechanisms.

\section{Methods} 

\subsection{Density map representation using 3D Gaussians}
We represent the protein density map with a group of learnable 3D Gaussians $\mathbb{G}^3=\{G^3_i\}_{i=1,\ldots,N}$ \cite[]{kerbl20233d}. Each Gaussian $\mathbb{G}^3_i$ depicts a local Gaussian-shaped density distribution, i.e.,
\begin{align}
    \mathbb{G}^3_i(\mathbf{x}|d_i,\mathbf{p}_i,\mathbf{\Sigma}_i)=d_i\cdot\exp(-\frac{1}{2}(\mathbf{x}-\mathbf{p}_i)^\mathsf{T}\mathbf{\Sigma}_i^\mathsf{T}(\mathbf{x}-\mathbf{p}_i)),
\end{align}
where $d_i$, $\mathbf{p}_i\in\mathbb{R}^3$, and $\mathbf{\Sigma}_i\in\mathbb{R}^{3\times3}$ define the density value, position, and covariance of a Gaussian, respectively. Note that protein molecules are composed of atomic spheres and are imaged at the same scale, we simplify the covariance matrix $\mathbf{\Sigma}_i$ into scalar deviation $s_i$ and identity matrix $\mathbf{I}$: $\mathbf{\Sigma}_i=(s_i\cdot \mathbf{I})(s_i\cdot \mathbf{I})^\mathsf{T}$. Thus, the target density at position $\mathbf{x}\in\mathbb{R}^3$ is the sum of the contributions from nearby isotropic Gaussians:
\begin{align}
    D(\mathbf{x})=\sum^N_{i=1}\mathbb{G}^3_i(\mathbf{x}|d_i,\mathbf{p}_i,\mathbf{\Sigma}_i).
\end{align}
Within 3D Gaussians, the compositional and conformational heterogeneity are directly modeled as changes in densities and positions, respectively. 

A good property of the Gaussian distribution is that its integral along one coordinate axis (usually $z$ row in the Cryo-EM domain) in 3D yields a 2D Gaussian distribution \cite[]{zha2024r, bae2024per}. The $\hat{\mathbf{x}}\in\mathbb{R}^2$, $\hat{\mathbf{p}}_i\in\mathbb{R}^2$ and $\hat{\mathbf{\Sigma}}_i\in\mathbb{R}^{2\times2}$ of projected Gaussians $\mathbb{G}^2_i$ are easily obtained by directly discarding the specific rows and columns of their corresponding 3D versions. The rendered particle projection $I_k\in\mathbb{R}^{H\times W}$ can be formulated as
\begin{align}
    I_k\in\mathbb{R}^{H\times W} &= \mathcal{R}(\mathbb{G}^3|\phi_k) \\
    &= C_k\circ P\circ T(\mathbb{G}^3|\phi_k) \\
    &= C_k\circ (\sum^N_{i=1}\mathbb{G}^2_i(\hat{d}_i,\hat{\mathbf{p}}_i,\hat{\mathbf{\Sigma}}_i)), \label{eq:projection}
\end{align}
where $\mathcal{R}$ denotes the complete rendering process, $C_k$ denotes the CTF modulator, $P$ is the simple projection operator along the $z$ row, and $T$ transforms the 3D Gaussians given pose $\phi_k$. For computational efficiency, $T$ actually calculates the inverse matrix of the particle pose to align the viewing direction with the coordinate axis. 

\subsection{Heterogeneous reconstruction network with Gaussian encoder}

Consistent with most existing work, the heterogeneous reconstruction network adopts an encoder–decoder architecture that maps raw particle images to their possible deformed volumes. Within the framework of 3D Gaussians, the deformation of flexible structures is characterized by the change of Gaussian parameters, whereas the invariant parameters correspond to the canonical density map obtained from homogeneous reconstruction. Our heterogeneous reconstruction network consists of three main modules, two encoders and one decoder, all implemented using multilayer perceptrons with Relu activation functions (see Figure \ref{fig:network} Supplementary Material section ``Network architecture''). First, the raw particle images are processed by an image encoder made up of three hidden layers, which transforms them into $i$-dimension latent variables. In neural radiance field-based methods, position encoding is introduced to facilitate the network determining particle poses and learning high-frequency features. Similarly, we design a Gaussian encoder to assign a unique identifier to each Gaussian, helping to distinguish and aggregate different Gaussians. This encoder takes the density, scaling, position, and rotation parameters of each Gaussian as input and produces a $j$-dimension latent embedding. This per-Gaussian strategy has no limit on the number of Gaussians, allowing the incorporation of given atomic models and yielding more accurate estimations.

The deformation decoder also follows the per-Gaussian paradigm, accepting the concatenation of an $i$-dimension latent variable and a $j$-dimension Gaussian embedding, and outputting predicted deformation for each Gaussian. This design allows flexible selection of Gaussians of interest and observation of their parameter variations, without re-estimating all Gaussians. The estimated changes are applied to the initial 3D Gaussians and then projected according to specified particle poses. The resulting projection of the deformed volume is subsequently transformed into the Fourier domain and corrupted by the CTF.

\subsection{Loss function with geometric constraints}

The heterogeneity information is naturally expressed in the real space, and so are 3D Gaussians. Any change in their parameters can be visualized. Given cryo-EM images $\mathbf{y}_k$, we define all loss functions on the real space, including reconstruction loss and geometric constraints. 

\textbf{Reconstruction Loss}. We employ $L$-1 loss to compare the difference between the projection $I'_k$ of deformed volume and raw images $\mathbf{y}_k$:
\begin{align}
    L_{rec}=\sum_k||I'_k-\mathbf{y}_k||_1^1=\sum_k||\mathcal{R}(\mathbb{G}^3+\Delta\mathbb{G}^3_k|\phi_k)-\mathbf{y}_k||_1^1,
\end{align}
where $\Delta\mathbb{G}^3_k$ represents the changes of Gaussian parameters.

\textbf{Geometric Constraints}. Overfitting can easily occur in the absence of well-designed regularization, particularly when dealing with high-noise experimental images. In such cases, Gaussians tend to aggregate, resulting in low-resolution structures that contribute uniformly across inputs. Considering the inherent continuity of protein motions, smoothness and local rigidity are essential properties to maintain. For this purpose, we impose two types of constraints on the estimated parameters: one for Gaussian embedding and the other just on the Gaussian displacement. For Gaussian embedding $\mathbf{j}_n$, given the average value of nearest point distance $d_{mean}$, we consider the neighboring Gaussians within a radius of $\mu_1=2.5\times d_{mean}$; 
\begin{align}
    L_{emb} &=\sum_n\sum_{s\in \mathcal{N}(n,\mu_1)}\omega_{n,s}||\mathbf{j}_n-\mathbf{j}_s||_2^2 \\
    \omega_{n,s} &= \exp{(-\lambda_\omega||d_{n,s}||_2^2)},
\end{align}
where $\mathcal{N}(n,\mu)$ denotes the set of Guassians within distance $\mu$ from Gaussian $n$, and $d_{n,s}$ is the initial space distance between Gaussians $n$ and $s$. The weight $\omega_{n,s}$ controls the strength of constraints. The closer the two Gaussians are, the stronger the constraint is. This embedding constraint encourages adjacent Gaussians to undergo similar parameter changes, thereby preserving local consistency in both compositional and conformational heterogeneity.
For the displacements of Gaussian, we first enforce the local rigidity of deformations by constraining the relative distance between adjacent Gaussians.
\begin{align}
    L_{geo}^1 &=\sum_n\sum_{s\in \mathcal{N}(n,\mu_2)}\omega_{n,s}||\hat{d}_{n,s}-d_{n,s}||_2^2,
\end{align}
where $\hat{d}_{n,s}$ denotes the distances between deformed Gaussians, and threshold $\mu_2$ is set to the $1.5\times d_{mean}$ in this paper. This strict distance constraint is applied only within a limited range; otherwise, deformation estimation tends to degenerate into missing-component estimation. We further impose an additional constraint on the orientations of the displacement vectors $\mathbf{v}_n$, which operates over a broader neighborhood of Gaussians:
\begin{align}
    L_{geo}^2 &=\sum_n\sum_{s\in \mathcal{N}(n,\mu_1)}\omega_{n,s}\cdot \cos{(\mathbf{v}_n, \mathbf{v}_s)}.
\end{align}

\section{Availability of data and materials}
The cryo-EM datasets and protein atomic models used in the paper are publicly available through EMPIAR and PDB, respectively.

\section{Funding}
We wish to thank Wenfei Li, Xiang Zhang, and Jianyi Yang for their insightful feedback and constructive criticism on the manuscript. This research was supported by the National Natural Science Foundation of China [W2511070, 624B2089, 32241027], the National Key Research and Development Program of China [2021YFF0704300], and the Natural Science Foundation of Shandong Province [ZR2023YQ057].


\bibliographystyle{natbib}
\bibliography{ref}

\end{document}


\title{Resolving compositional and conformational heterogeneity in cryo-EM with deformable 3D Gaussian representations}
\renewcommand\footnotemark{}
\author{Bintao He$^{1,\dagger}$, Yiran Cheng$^{1,2,\dagger}$, Hongjia Li$^{3}$, Xiang Gao$^{4}$, Xin Gao$^{5,*}$, Fa Zhang$^{3,*}$,  Renmin Han$^{1,2,*}$  \thanks{$^{\dagger}$These authors should be regarded as Joint First Authors; $^*$All correspondence should be addressed to Xin Gao (xin.gao@kaust.edu.sa), Fa Zhang (zhangfa@bit.edu.cn) and Renmin Han (hanrenmin@sdu.edu.cn).}}
\date{\small $^1$Research Center for Mathematics and Interdisciplinary Sciences (Ministry of Education Frontiers Science Center for Nonlinear Expectations), Shandong University, Qingdao 266237, China; $^2$College of Medical Information and Engineering, Ningxia Medical University, Yinchuan 750004, China;$^3$School of Medical Technology, Beijing Institute of Technology, Beijing 100081, China; $^4$State Key Laboratory of Microbial Technology, Shandong University, Qingdao 266237, China; $^5$King Abdullah University of Science and Technology (KAUST), Computational Bioscience Research Center (CBRC), Computer, Electrical and Mathematical Sciences and Engineering (CEMSE) Division, Thuwal, 23955, Saudi Arabia.}
\maketitle
\section{IgG simulation}

\begin{figure}[H]
	\centering
	\includegraphics[width=0.8 \textwidth]{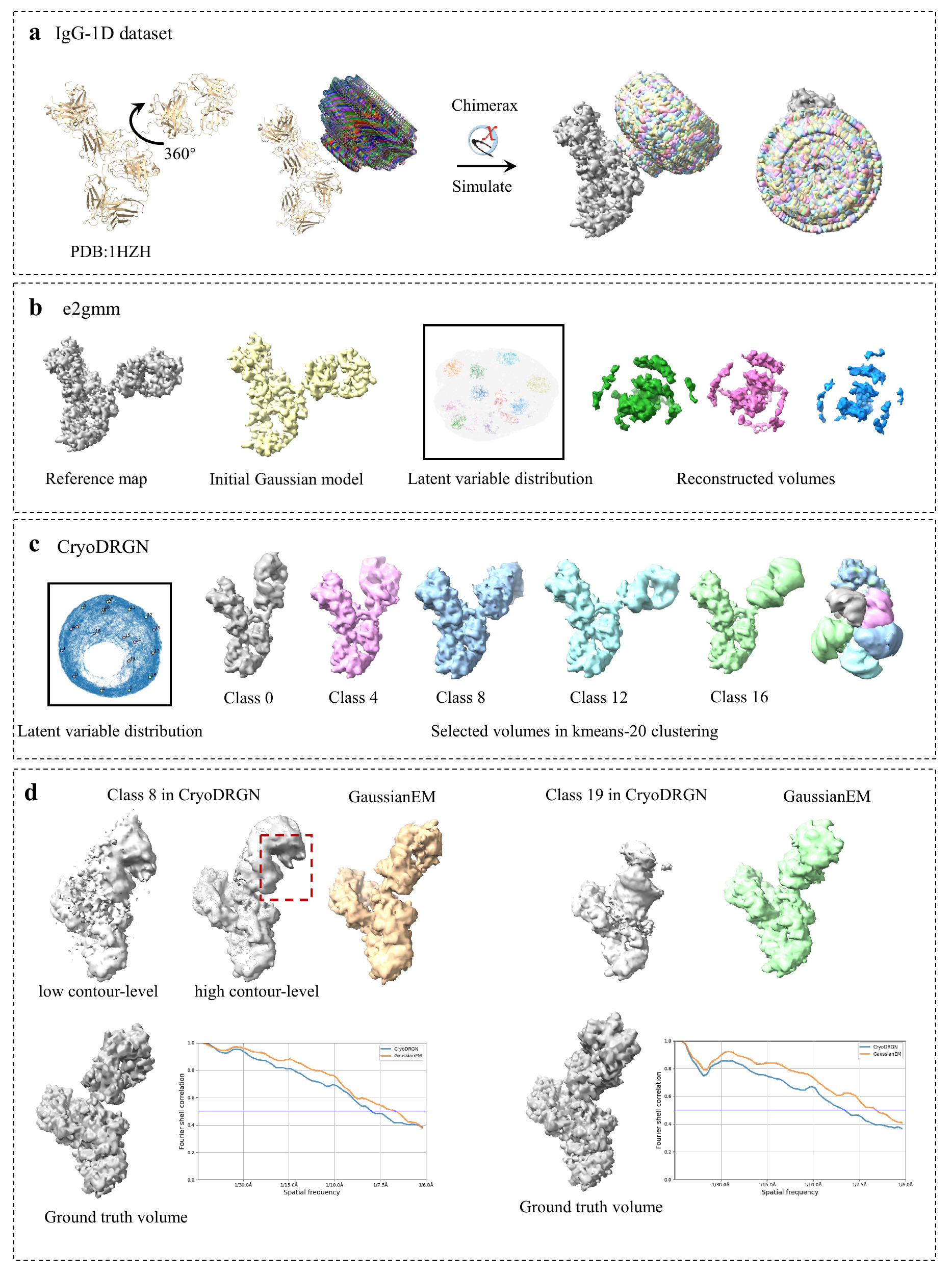}
	\caption{\textbf{a} Overview of IgG-1D dataset. \textbf{b} Heterogeneity analysis of e2gmm, which fails to reconstruct physically reasonable structures. \textbf{c} Heterogeneity analysis of CryoDRGN with 20 clusters. \textbf{d} Two representative examples where CryoDRGN produces inconsistent local structures, with the corresponding plausible reconstructions from GaussianEM shown on the right.}
	\label{fig:IgG}
\end{figure}

The conformational heterogeneity in the IgG-1D dataset is produced by rotating a dihedral angle connecting one of the fragment antibody (Fab) domains, simulating a simple one-dimensional continuous circular motion \cite[]{jeon2024cryobench}. Although we provided a well-designed initial model, Gaussian-based method e2gmm failed to reconstruct reasonable structures under the large-scale motion. Voxel-based approach CryoDRGN successfully captured the overall circular rotation of the Fab domains, but occasional structural loss occurred in localized areas, as shown in Figure \ref{fig:IgG}d.

\begin{figure}[H]
	\centering
	\includegraphics[width= 0.8\textwidth]{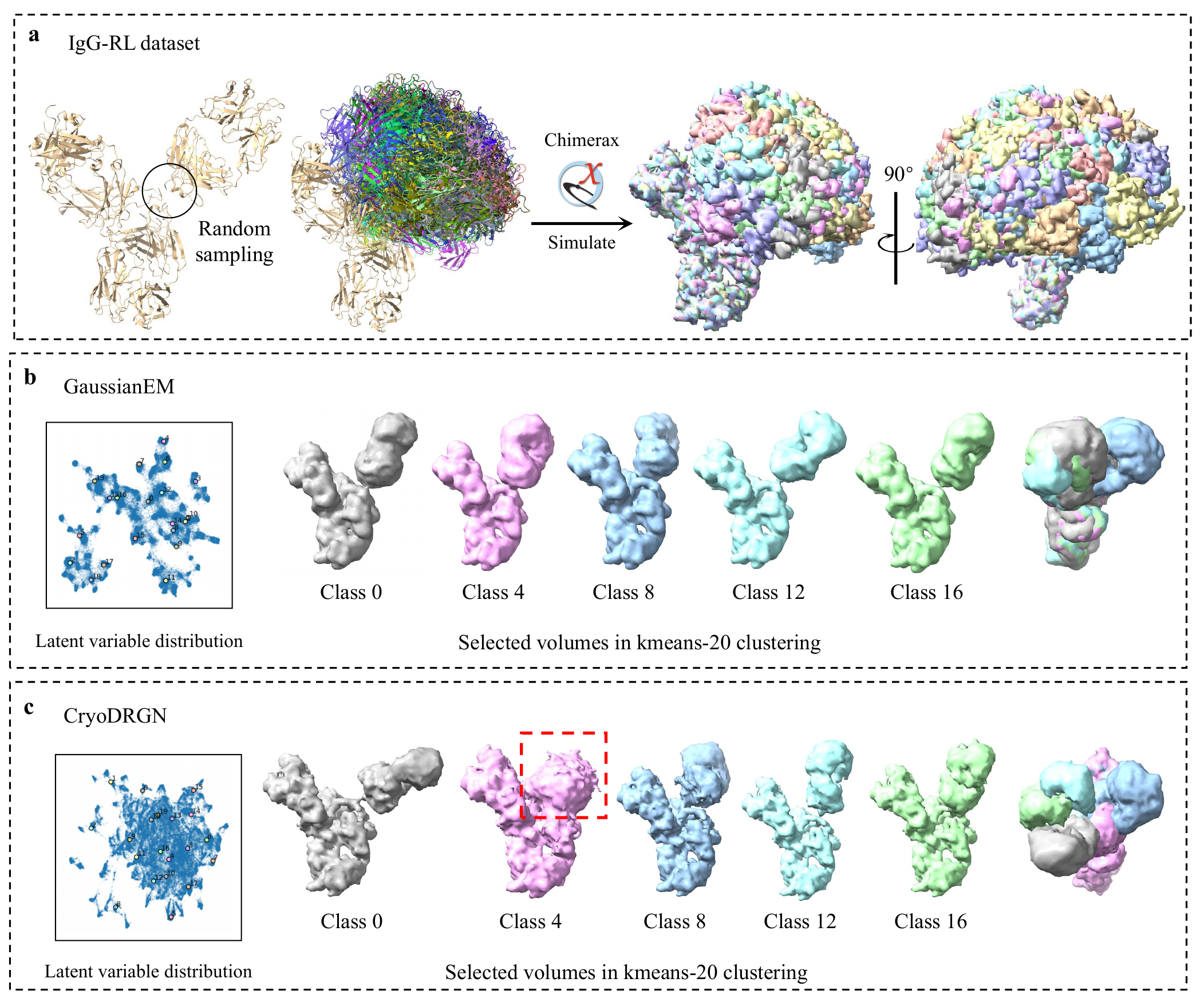}
	\caption{\textbf{a} Overview of IgG-RL dataset. \textbf{b} Heterogeneity analysis of GaussianEM with 20 clusters. \textbf{c} Heterogeneity analysis of CryoDRGN with 20 clusters. }
	\label{fig:IgG_RL}
\end{figure}

By incorporating randomly sampled linker conformations, the IgG-RL dataset more realistically captures conformational dynamics than IgG-1D. We present the latent conformation space and heterogeneous reconstruction results of GaussianEM and CryoDRGN on the IgG-RL dataset in Figure \ref{fig:IgG_RL}b and Figure \ref{fig:IgG_RL}c. The latent conformation map of GaussianEM exhibits a more distinct separation of particle clusters than that of CryoDRGN, and the reconstruction results of CryoDRGN suffer from missing densities in localized regions, as highlighted by the red dashed box.

\section{Pre-catalytic spliceosome}

\begin{figure}[H]
	\centering
	\includegraphics[width=0.8 \textwidth]{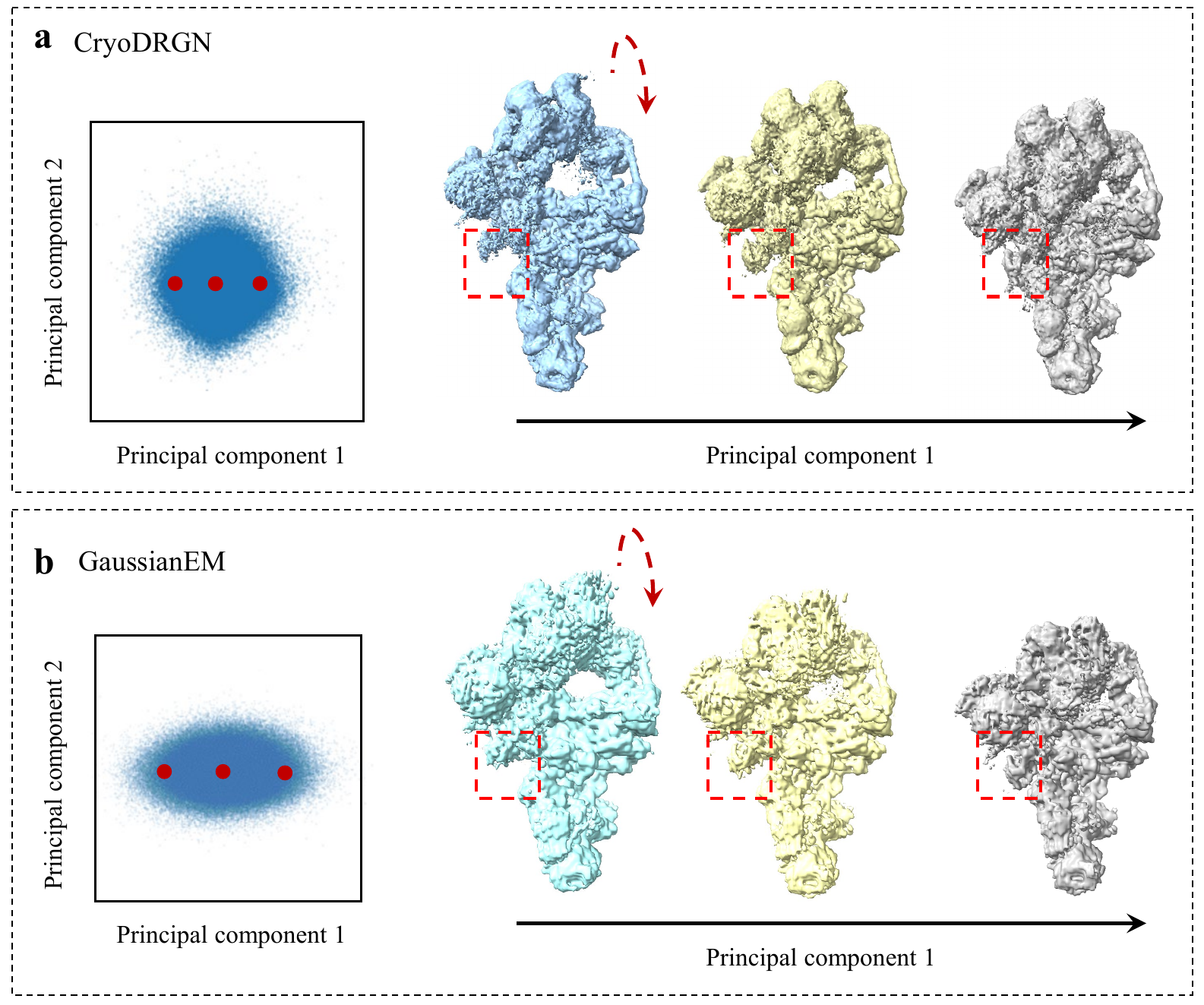}
	\caption{\textbf{Reconstructed maps along the first principal component.} \textbf{a} Density maps reconstructed by CryoDRGN. \textbf{b} Density maps reconstructed by GaussianEM.}
	\label{fig:10180_drgn}
\end{figure}

In conformational heterogeneity analysis, voxel-based methods may introduce artifacts. When generating maps of continuous motion, local structural changes can sometimes become incoherent. As shown in Figure \ref{fig:10180_drgn}a, CryoDRGN \cite[]{zhong2021cryodrgn} reconstructed three states along the first principal component. The region highlighted by the dashed red box exhibits a downward displacement but is structurally inconsistent. In contrast, our method estimates the deformation applied to the consensus model, preserving the local structure and clearly capturing the overall moving trend.

\section{T6SS effectors}

\begin{figure}[H]
	\centering
	\includegraphics[width=0.8 \textwidth]{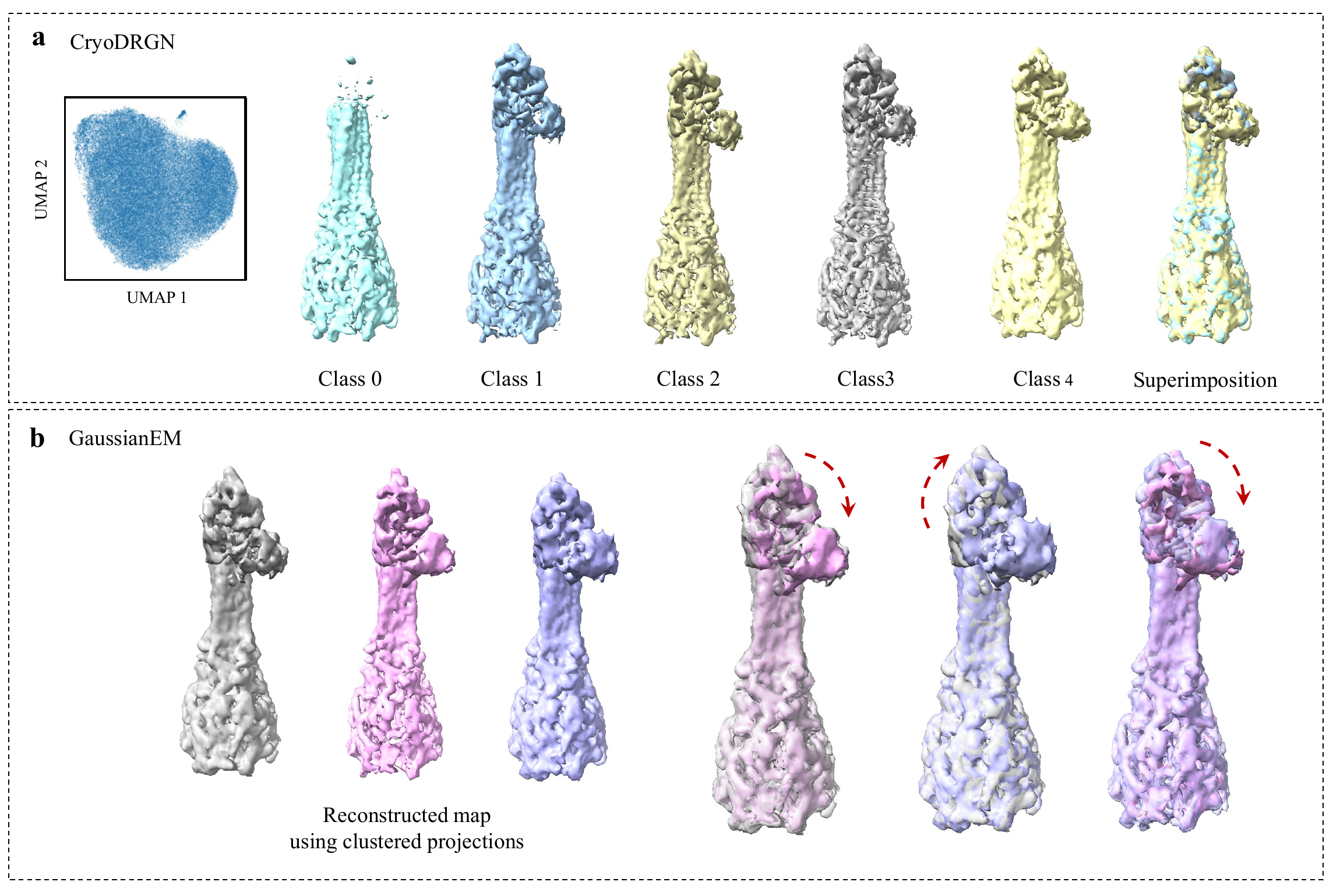}
	\caption{\textbf{a} Five discrete state maps reconstructed by CryoDRGN. The rightmost image shows the superposition of these five volumes. \textbf{b} Density maps reconstructed in CryoSPARC using clustered images from GaussianEM. The superposition of any two maps highlights the motion of the effector region.}
	\label{fig:gao_drgn}
\end{figure}

CryoDRGN was applied to the T6SS effector dataset, and five representative maps were reconstructed by clustering. CryoDRGN successfully captured the absence of effectors, but it failed to reproduce the additional strip structure described in the manuscript section “T6SS effectors”. When the five reconstructed maps were superimposed, no clear displacement was observed in the effector region (Figure \ref{fig:gao_drgn}a). In contrast, GaussianEM effectively captured the motion of effectors. The clustered images were fed into CryoSPARC \cite[]{punjani2017cryosparc} to perform homogeneous reconstruction. The superimposition of any two maps exhibits slight movements of the effector region, demonstrating the strong ability of GaussianEM to model both the composition and conformational heterogeneity.

\section{$\alpha V\beta8$ integrin bound to latent TGF-$\beta$}

\begin{figure}[H]
	\centering
	\includegraphics[width=0.8 \textwidth]{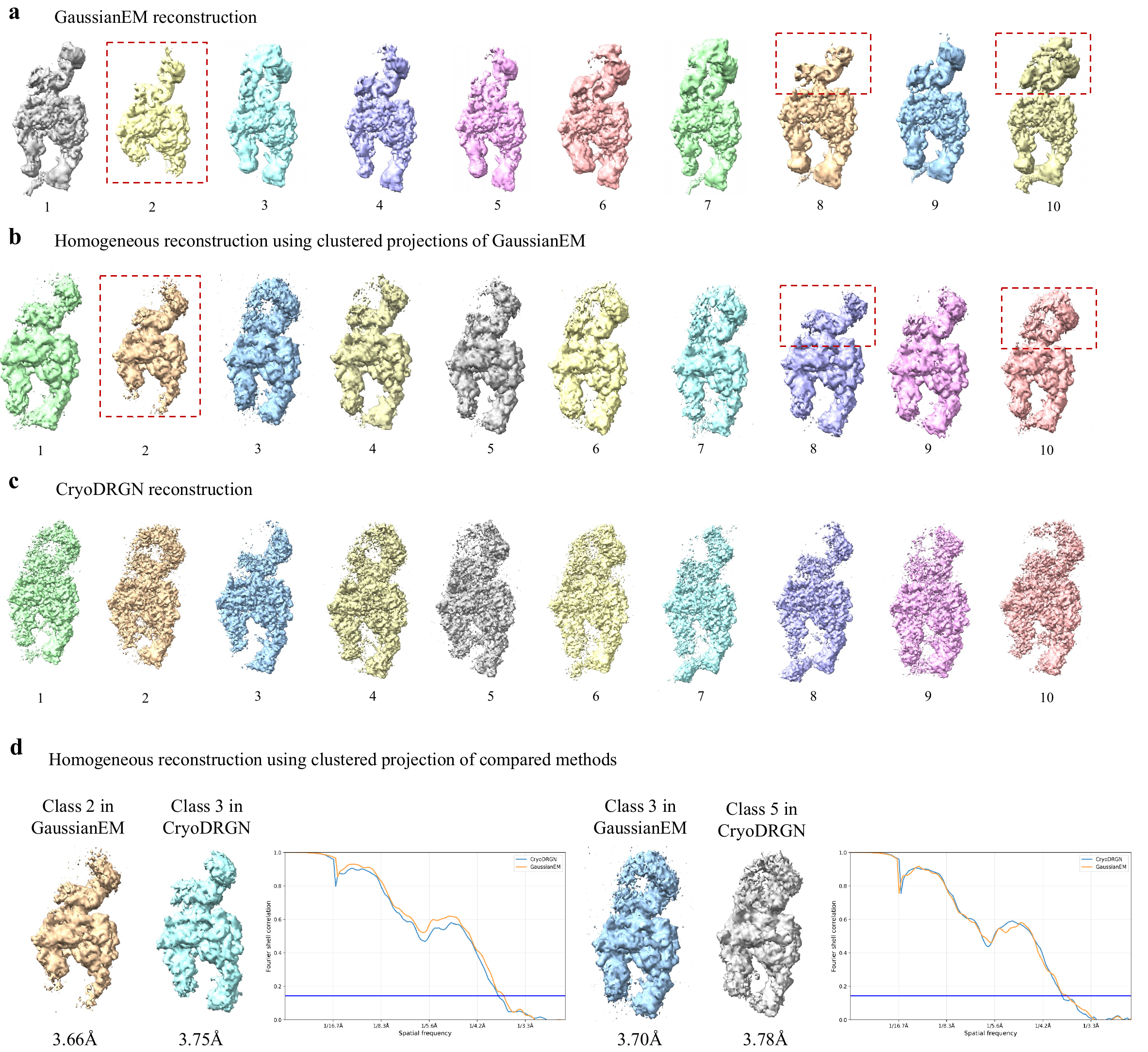}
	\caption{Ten representative density maps reconstructed by GaussianEM (\textbf{a}), CryoSPARC (\textbf{b}) and CryoDRGN (\textbf{c}). \textbf{d} The homogeneous reconstruction of manually selected clusters from GaussianEM and CryoDRGN, that are presumed to correspond to one another.}
	\label{fig:10343}
\end{figure}

In addition to $\alpha V\beta8$ integrin (EMPIAR-10345) described in the main text, we also performed GaussianEM on the $\alpha V\beta8$ integrin bound to latent TGF-$\beta$ (EMPIAR-10343). This dataset consists of 68356 particle images with a box size of 300 pixels of width 1.345 \AA. Figure \ref{fig:10343} presents ten representative density maps reconstructed by GaussianEM and CryoDRGN. GaussianEM identified a broader range of possible conformations, highlighted by dashed red boxes. These conformations were further validated through homogeneous reconstruction of clustered projections (Figure \ref{fig:10343}b). Figure \ref{fig:10343}d presents the homogeneous reconstructions of manually selected clusters from GaussianEM and CryoDRGN, which are presumed to correspond to one another. The higher resolution of the maps from GaussianEM suggests that its projection clustering may be more compact.

\section{Network architecture}

\begin{figure}[H]
	\centering
	\includegraphics[width=1.0 \textwidth]{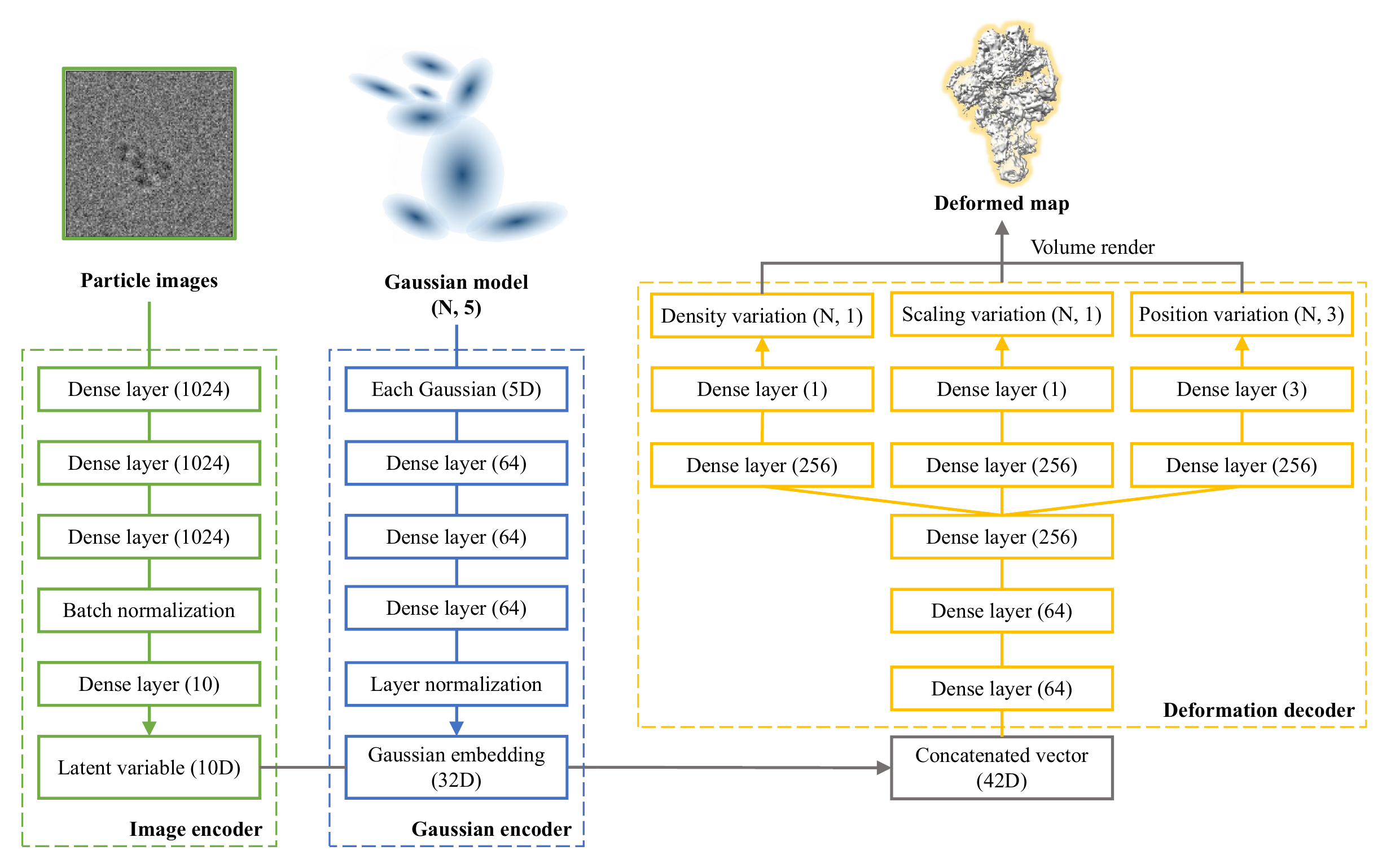}
	\caption{Detailed structure of the default network.}
	\label{fig:network}
\end{figure}

Figure \ref{fig:network} shows the detailed network architecture used for the examples shown in the main text. The image encoder processes particle images and projects them into a latent conformational space. The Gaussian encoder assigns a unique identifier to each Gaussian component, enabling the network to recognize and localize individual Gaussians. The latent variable from the image encoder and the Gaussian embedding from the Gaussian encoder are concatenated and passed to the deformation decoder, which independently estimates variations in density, scaling, and position.

\bibliographystyle{natbib}
\bibliography{ref}
\newpage